\documentclass[lettersize,journal,final]{IEEEtran}
\usepackage{amsmath,amsfonts}
\usepackage{algorithmic}
\usepackage{algorithm}
\usepackage{array}
\usepackage[caption=false,font=normalsize,labelfont=sf,textfont=sf]{subfig}
\usepackage{textcomp}
\usepackage{stfloats}
\usepackage{url}
\usepackage{verbatim}
\usepackage{graphicx}
\usepackage{cite}
\usepackage{subfloat}
\usepackage{float}
\usepackage{hyperref}
\usepackage{tikz,xcolor,hyperref}% Make Orcid icon
\newtheorem{proof}{Proof}[section]

\newtheorem{definition}{Definition}  
\definecolor{lime}{HTML}{A6CE39}
\DeclareRobustCommand{\orcidicon}{%
	\begin{tikzpicture}
		\draw[lime, fill=lime] (0,0) 
		circle [radius=0.16] 
		node[white] {{\fontfamily{qag}\selectfont \tiny ID}};    \draw[white, fill=white] (-0.0625,0.095) 
		circle [radius=0.007];    \end{tikzpicture}
	\hspace{-2mm}}
\foreach \x in {A, ..., Z}{%
	\expandafter\xdef\csname orcid\x\endcsname{\noexpand\href{https://orcid.org/\csname orcidauthor\x\endcsname}{\noexpand\orcidicon}}
}
\begin{document}

\title{GSpect: Spectral Filtering for Cross-Scale Graph Classification}

\author{
	\IEEEauthorblockN{
		Xiaoyu Zhang,
		Wenchuan Yang,
		Jiawei Feng, 
		Bitao Dai, 
		Tianci Bu, and
		Xin Lu\orcidB{}}
%	\\
	%\IEEEauthorblockA{\IEEEauthorrefmark{1}College of Systems Engineering, National University of Defense Technology, Changsha 410073, Hunan, PR China}
	%\\
	%\IEEEauthorblockA{\href{mailto:xiaoyu_zhang2023@163.com}{xiaoyu\_zhang2023}@163.com, \href{mailto:wenchuanyang97@163.com}{wenchuanyang97}@163.com,
		%\href{fengjiawei126@gmail.com}{fengjiawei126}@gmail.com,
		%\href{daibitao@nudt.edu.cn}{daibitao}@nudt.edu.cn,
		%\href{btc010001@gmail.com}{btc010001}@gmail.com,
		%\href{xin.lu.lab@outlook.com}{xin.lu.lab}@outlook.com}% 给人名附上邮箱地址

\thanks{This work was supported by the National Natural Science Foundation of China (72025405, 72088101), the National Social Science Foundation of China (22ZDA102), the Hunan Science and Technology Plan Project (2020TP1013, 2020JJ4673, 2023JJ40685), the Shenzhen Basic Research Project for Development of Science and Technology (202008291726500001), and the Innovation Team Project of Colleges in Guangdong Province (2020KCXTD040). The authors declare that they have no conflict of interest. (Corresponding author: Xin Lu.)
\\
Email addresses: xiaoyuzhang\_2023@163.com (Xiaoyu Zhang),
wenchuanyang97@163.com (Wenchuan Yang),
fengjiawei126@gmail.com (Jiawei Feng), daibitao@nudt.edu.cn
(Bitao Dai), btc010001@gmail.com (Tianci Bu),
xin.lu.lab@outlook.com (Xin Lu)}}
%	\IEEEauthorblockA{Corresponding Author: Xin Lu \quad Email: xin.lu.lab@outlook.com}}

        % <-this % stops a space
%\thanks{This paper was produced by the Group of Big Data and Complex Systems.}% <-this % stops a space

%\markboth{IEEE TRANSACTIONS ON NETWORK SCIENCE AND ENGINEERING}
\markboth{}{X. Zhang et al.: GSpect: Spectral Filtering for Cross-scale Graph Classification}
%\markboth{2}{1}
%\markright{X. Zhang et al.: GSpect: Spectral Filtering for Cross-scale Graph Classification}


\maketitle

\begin{abstract}
Identifying structures in common forms the basis for networked systems design and optimization. However, real structures represented by graphs are often of varying sizes, leading to the low accuracy of traditional graph classification methods. These graphs are called cross-scale graphs. To overcome this limitation, in this study, we propose GSpect, an advanced spectral graph filtering model for cross-scale graph classification tasks. Compared with other methods, we use graph wavelet neural networks for the convolution layer of the model, which aggregates multi-scale messages to generate graph representations. We design a spectral-pooling layer which aggregates nodes to one node to reduce the cross-scale graphs to the same size. We collect and construct the cross-scale benchmark data set, MSG (Multi Scale Graphs). Experiments reveal that, on open data sets, GSpect improves the performance of classification accuracy by 1.62\% on average, and for a maximum of 3.33\% on PROTEINS. On MSG, GSpect improves the performance of classification accuracy by 15.55\% on average. GSpect fills the gap in cross-scale graph classification studies and has potential to provide assistance in application research like diagnosis of brain disease by predicting the brain network's label and developing new drugs with molecular structures learned from their counterparts in other systems.  
\end{abstract}
\section{INTRODUCTION}

\IEEEPARstart{D}{ata} that have a non-Euclid structure\textemdash such as protein structures\cite{whitford2013proteins}, social networks\cite{lu2016detecting} and compounds\cite{schadt2009network}\textemdash are often represented by graphs with nodes and edges. As structure determines function in many networked systems, graph classification (for exact definition, please refer to \ref{def}) is a fundamental research problem in numerous fields. For example, in computer vision, graph classification methods are used to measure the similarity of human action recognition among graphs\cite{wu2014human}. In neuroscience, researchers use graph classification methods to study the similarity of brain networks\cite{lee2020deep}. In chemistry, graph classification methods are used to learn the similarity of chemical compounds in terms of their effect on reaction partners\cite{brown2009chemoinformatics}. Fields such as bioinformatics and molecular chemistry often encounter a problem named graph classification: graphs with different structures possess totally different functions. Researchers must separate graphs with different structures to select appropriate graphs in a short time. For example, Alzheimer's disease (AD) is known to be caused by structural changes in the brain. Researchers take samples of brain networks and determine whether the sample is likely to develop AD\cite{wen2020convolutional}.  %So it's a very significant problem to find technologies with similar structures and specific functions, and graph classification methods can solve this problem.

Researchers have proposed numerous methods to accomplish the graph classification problem, like graph kernels\cite{nikolentzos2021graph}. Graph kernels can be used for graph classification. However, these methods often define graphs in a heuristic manner, thereby resulting in low explainability and flexibility of these methods. It is for this reason that graph neural networks (GNNs) have become a popular method for graph classification tasks in recent years due to their ability to learn node and edge representations and capture messages from complex graph structures. Researchers usually design a GNN convolution layer to obtain the graph representation and design a GNN-based pooling layer to reduce the size of the graphs. One of the most classic definition of GNN convolution layers is spectral-based GNN. %, whose creators drew inspiration from signal processing\cite{shuman2013emerging}. The spectral-based GNN is defined as:
%\begin{equation}
%	H^{k}  = \sigma  (\sum U\Omega^{k}  U^{T}H^{k-1})
%\end{equation}
%Where $H^{k-1}$ is the input signal, $H^{k}$ is the output signal, $\Omega^{k}$ is a diagonal matrix which filled with learnable parameters and $ \sigma ()$ is an activating function. 
Spectral-based GNN use diagonal spectral filters to capture messages on the spectrum. This method has been wildly used for node classification and edge prediction tasks\cite{kipf2016semi}\cite{zhuang2018dual}. However, spectral-based methods have a few limitations\cite{wu2020comprehensive}: First, any perturbation to the graph results in the change of the graph's eigenvalues. Second, the learned filters are size-dependent. One graph determines a unique network structure, which implies that it is difficult to be applied to graphs with different sizes. So it is difficult to be used in the cross-scale graph classification tasks. 

Traditional graph classification methods only work on comparing structure of similar sizes\cite{dobson2003distinguishing}\cite{yanardag2015deep}\cite{debnath1991structure}. However, in practice, structures of an order-of-magnitude difference in size may have the same function. For example, in biology, the structure of proteins, which possesses critical functions\textemdash such as immune signaling\cite{silva2019novo}, targeted therapeutics\cite{cao2020novo}, sense-response systems\cite{glasgow2019computational} and self-assembly materials\cite{hsia2016design}\textemdash can be represented as graphs whose nodes represent the atoms and edges represent the chemical bonds. The protein structure determines its function. Certain proteins which have the same functions usually have similar graph structures. However, these protein-graphs occasionally have an order-of-magnitude difference in the number of nodes\cite{redfern2008exploring}. This set of graphs is called cross-scale graphs. Cross-scale graph classification tasks refers to dividing the graphs with an order-of-magnitude difference in the number of nodes into sets. Research on cross-scale graphs is an important research direction in the field of complex networks. Cross-scale graphs plays an important role in the practical applications such as network clustering\cite{du2010clustering}, hierarchical reduction\cite{tan2003general}, and state partition\cite{slota2016complex}. Researchers require cross-scale graph classification algorithms to select structure-similar but cross-scale proteins from a huge selection space. \cite{liu2022size} have proposed methods tailored to datasets of varying graph sizes, but these studies have exclusively designed methods for small-scale, sparse graphs (don't consider large-scale graphs) and conducted experiments only on publicly available datasets with similar graph sizes. There is no available method for cross-scale graphs' classification task and there are no open data sets with graphs which have enough difference (up to $10^3$) in the number of nodes. 

Graph Wavelet Transform (GWT) is a powerful tool for capturing multi-scale graph representations\cite{hammond2011wavelets}\cite{shuman2013emerging} due to its unique properties, making it becomes a powerful tool to solve cross-scale graph classification problems. The advantages of GWT is listed as follows. GWT offers multi-scale analysis capabilities, effectively representing both local and global features of graph structures\cite{hammond2011wavelets}. Besides, its localization properties in both spatial and frequency domains enable efficient capture of local structural information\cite{defferrard2016convolutional}. In addition, GWT typically produces sparse representations of graph signals, facilitating key feature extraction\cite{tremblay2014graph}. Compared to global spectral methods, GWT often demonstrates higher computational efficiency, especially for large-scale graphs\cite{xu2019graph}. Apart from that, it naturally adapts to irregular graph structures, a challenge for traditional wavelet transforms\cite{coifman2006diffusion}. Recently, it is proved that GWT allows for cross-scale information integration, helping to capture hierarchical structures in graphs\cite{donnat2018learning}. Moreover, it exhibits robustness to minor structural changes, which is valuable when dealing with noisy data \cite{shuman2013emerging}. These characteristics make GWT a versatile and effective tool for multi-scale graph representation, with wide applications in graph classification, node classification, and graph signal processing.

In this article, we modified the spectral-based GNN using the graph wavelet theory and design a novel framework (GSpect) to accomplish cross-scale graph classification tasks. Specifically, considering the characteristic that the wavelet function can accurately capture the signal information in different frequency bands, we first use a graph wavelet neural network as the convolution layer for graph classification tasks. %Different from extant graph wavelet structure, we use non-sparse learnable filters (rather than diagonal matrix) to aggregate multi-scale graph signals to get global graph representations. Compared with other graph wavelet neural network, this improvement leads to the mixture of nodes' representation, which contributes to generate graph-level representation from graphs.  
%Compared with the traditional spectral-based GNN, the spectral graph wavelet transform has the advantage of capturing multi-scale messages and observe the graph signal step by step from coarse to fine.
Second, we design a graph pooling layer. Compared with other spectral clustering methods\cite{bruna2013spectral}\cite{defferrard2016convolutional}, we directly perform Fourier transformation on the graph's adjacency matrix and node attributes directly to obtain the frequency domain representation. We use spectral filters to filter high-frequency information and resize the graph on the principle of the save-most message. Third, considering the fact that there are no  appropriate cross-scale graph classification data sets, we collect three classes of empirical networks---covering the set of protein structure data, macromolecular compound structure data, and social networks in combination with the three typical modeled networks of ER\cite{erdHos1960evolution}, WS\cite{watts1998collective} and BA\cite{barabasi1999emergence}, to create a synthesis cross-scale graph classification benchmark data set MSG. We verify the performance of GSpect both on the open data sets and on MSG. 

This article makes the following contributions: 

1. We apply the graph wavelet theory to graph classification tasks and use the graph wavelet convolution layer to aggregate multi-scale information from graphs and generate graph-level representation. 

2. We design a pooling layer by using non-square learnable filters in the frequency domain to filter unusable messages and generate a low-order graph (refers to a graph structure obtained by aggregating or filtering nodes, resulting in a structure with fewer nodes and edges). 

3. We collect cross-scale graph data and generate a cross-scale graph data set MSG and conduct experiments on both open data sets and MSG. We test the classification accuracy and the results indicate that on open data sets, GSpect improves the performance of classification accuracy by 1.62\% on average, and for a maximum of 3.33\% on PROTEINS. On MSG, GSpect improves the performance of classification accuracy by 15.55\% on average of all state-of-art comparative models. 

The remainder of this article is organized in the following manner: Section II presents the related works. Section III proposes GSpect in detail. Section IV presents the experimental results, including the comparison experiment, ablation study, and sensitivity analysis. Section V summarizes our contributions and future directions. 

\section{RELATED WORKS}
\subsection{Graph Kernel Models for Graph Classification}
Graph kernels capture the similarity between graphs for graph classification tasks. Given a set of graphs, the graph kernel methods aim to learn the kernel function that captures the similarity between any two graphs. Traditional
graph kernels, such as random walk kernel, subtree kernel, and shortest-path kernels are widely used in graph classification tasks\cite{nikolentzos2021graph}\cite{ma2021deep}. The WL algorithm\cite{shervashidze2011weisfeiler} maps the original graph to a sequence of graphs whose node attributes are generated from graph topology and label information. A kernel family, including an efficient kernel family of comparison subtree patterns, can be defined from this WL sequence. This algorithm has became one of the most widely used graph kernel methods for graph classification. %Yanardag et al.\cite{yanardag2015deep} proposed one deep graph kernel approach. For a given set of graphs, they decomposed each graph into its substructures. Next, they regard the substructure as a word in the CBOW (continuous bag of words). Skip-gram is used to learn the potential representation measure co-occurrence of substructures from the graphs. 
Al-Rfou et al.\cite{al2019ddgk} proposed deep divergence graph kernels (DDGK). DDGK learn kernel functions for a pair of graphs. Given two graphs $G_1 $ and $ G_2 $, this method learns a kernel function $ K(.) $ as a similarity metric function for graphs. The function is defined in the following manner:
\begin{equation}
	k\left(G_{1},G_{2}\right)=\left\|\Psi\left(G_{1}\right)-\Psi\left(G_{2}\right)\right\|^{2}
	\label{kernel},
\end{equation}
where $\Psi\left(G_{1}\right)$ is the graph representation of $G_1$. This method learn the graph representation by computing the divergence of the target graph. Given a set of source graphs $ { G_1, G_2, ... , G_N } $, a graph encoder is the representation of each graph in the set. Then, for the target graph $ G_i $, the divergence between $ G_i $ and the source graph is computed to measure the similarity. The equation of divergence between $G_a$ and $G_b$ is as given bellow: 
\begin{equation}
	\mathcal{D}^{\prime}\left(G_{a} \| G_{b}\right)=\sum_{v_{i} \in V_{a}} \sum_{j , e_{i j} \in E_{a}}-\log \operatorname{Pr}\left(v_{j} \mid v_{i}, H_{b}\right),
\end{equation}
where $a$ is the encoder trained on graph $G_a$. $D'(G_a \parallel G_b$ represents the divergence from graph $G_a$ to graph $G_b$. $\Pr(v_j | v_i, H_b)$ represents the probability of node $v_j$ occurring given node $v_i$ under the encoder $H_b$ of graph $G_b$.
%Meanwhile, there are many other graph kernel methods for graph classification, like \cite{du2019graph}\cite{tian2019rethinking}\cite{arora2019exact}. 

However, graph kernel models have a few limitations: Most of them have low computational efficiency, and graph kernel methods use kernel functions(like Equation \ref{kernel}) to measure the similarity between two graphs, which implies that graph kernel methods can't be used to handle graph classification problems with a lot of graphs. 
\subsection{Classic GNN Models for Graph Classification}
In recent years, researchers have become increasingly interested in the extension of the deep learning method to graphs. Driven by the success of deep neural networks, the researchers drew on the ideas of convolutional neural networks, recurrent neural networks, and auto encoder to define and design a neural network structure for processing graph data. Consequently, a new method called GNNs emerged. Researchers have designed a few GNN-based graph classification methods. For example, the graph convolutional network (GCN)\cite{kipf2016semi} is one of the earliest methods in this discipline. GCNs learn node representations and propagate them to other nodes using a spectral graph convolution technique. In many graph classification tasks, GCN have demonstrated state-of-the-art performance. However, GCNs have limitations in capturing long-range relationships and higher-order graph structures. To solve these problems, MPNN\cite{gilmer2017neural} applies a message-passing algorithm to learn node representations on the local graph structure. It has been demonstrated that MPNN are efficient in capturing higher-order graph topology and long-range relationships. With the development of the attention mechanism, graph attention networks (GATs)\cite{velivckovic2017graph} have become a popular method for graph classification. GATs use self-attention to learn node representations, thereby enabling the model to focus on only the key nodes in the graph. GATs have been shown to achieve state-of-the-art performance on many graph classification tasks. In addition to these methods, several other GNN variants have been proposed, such as graph isomorphism networks (GINs)\cite{xu2018powerful}. These techniques have improved the classification accuracy for a variety of graph classification problems.

As the size of graphs to be classified are usually different and cannot be directly compared, many methods apply graph pooling to resize the graphs to a unique size before the classification. A number of intuitive methods are used for graph pooling. For example, max-pooling and mean-pooling use the maximum or average value of a group of nodes to represent them \cite{duvenaud2015convolutional}. However, these methods lack flexibility, which reduces the competitiveness of these methods. To overcome these limitations, %Lee et al.\cite{lee2019self} propose a graph pooling method based on attention mechanism which allows the model selecting the important nodes in the graph. 
Ma et al.\cite{ma2019graph} introduced EigenPooling, an innovative approach rooted in the graph Fourier transform. This method leverages the spectral domain to effectively pool nodes in a graph. However, the pooling process is heuristic and cannot be optimised by machine learning algorithms. %Researchers\cite{gao2021ipool}\cite{gao2021topology} design different node drop pooling rules based on different evaluate functions. 
For this problem, currently available spectral clustering (SC) methods \cite{bruna2013spectral}\cite{bianchi2020spectral} were proposed to identify clusters, which are subsets of nodes that are more densely connected to each other than to the rest of the graph. However, it leads to more computational complexity. However, these methods only execute pooling once, which occasionally leads to the loss of key nodes. To solve this problem, Ying et al.\cite{ying2018hierarchical} developed the hierarchical pooling approach (DiffPool). They created the concept assign matrix that maps a set of nodes to a single node using GNN models. The function of the assignment matrix is given below:
\begin{equation}
	S^{(k)} = softmax[GNN_{k,pooling}(A^{(k)}, X^{(k)})],
\end{equation}
where $A^{(k)}$ and $X^{(k)}$ are the graph's adjacency matrix and graph representation matrix. $ GNN_{k,pooling} $ is a learnable function. In practice, DiffPool combines its pooling method with the differentiable graph encoder to make the architecture top-to-end trainable.

%Researchers have proposed many summaries for graph classification tasks. For example, Ma et al.\cite{ma2021deep} provide a comprehensive survey of the existing literature of graph similarity learning problem and discuss the challenges and future works of the problem. Liu et al.\cite{liu2022graph} provide a competitive review of graph pooling for graph neural networks. They proposed a taxonomy of existing graph pooling models and introduced each class of methods in details. Researchers can read these articles to get more information. %For the fairness problem in machine learing, Errica et al.\cite{errica2019fair} propose reevaluate popular models on open benchmarks. This work contributes to the fairness development of graph classification tasks. 
\subsection{Wavelet Transform-Based Research}

%Wavelet transform is a powerful cross-scale analysis tool. It represents signals as a combination of multiple localization, shift, and scaling bases (wavelet bases)\cite{daubechies1992ten}. % It builds on the concept of the short-time Fourier transform and overcome the drawback of the window size not changing with frequency. It can provide a time-frequency window that change with frequency, making it a useful tool for time-frequency analysis and processing of signals. 
%Its key characteristics are its ability to fully highlight certain aspects of the issue through transformation, its ability to localize time-frequency signal analysis, and its ability to gradually refine the signal (or function) at scale translation processes. 
%It achieves frequency subdivision at low frequency and time subdivision at high frequency, enabling automatic adaptation to the requirements of time-frequency signal analysis and can focus on any specifics of the signal, making up for the deficiency of Fourier transform. Spectral wavelet transform is the projection of graph signal from vertex domain to spectral domain using a set of wavelet bases. 

As a part of the spectral theory, the wavelet theory has been widely used in the field of image processing and signal analysis. For example, Yahia et al.\cite{yahia2022wavelet} use wavelet neural networks for image classification and attain high accuracy. 

Some researchers applied wavelet transform to the spectral graph theory. For example, Hammond et al.\cite{hammond2011wavelets} defined a wavelet function to project the graph to the wavelet domain, the equation is defined in the following manner:
\begin{equation}
	\psi_{f, i}(j)=\sum_{t=1}^{N} g\left(f \lambda_{t}\right) u_{t}^{\star }(i) u_{t}(j),
\end{equation}
where $ N $ is the number of vertices, $ \lambda _t $ is the $ t$-th eigenvalue of the graph Laplacian matrix, $ u_t $ is the eigenvector of the Laplacian matrix. The symbol $ \star $ denotes the complex conjugate operator, and $ g $ is the spectral graph wavelet generating kernel. This research is the first to propose the concept of the graph wavelet transform. However, it does not combine graph wavelet theory and deep learning. 

%Graph wavelet transform is gradually used in the design of GNNs. For graph representation learning region, researchers have designed graph wavelet neural networks\cite{xu2019graph} for node classification tasks and get satisfactory results. Graph wavelet neural networks have a more formal mathematical expression and can capture multi-scale messages from graph structures and have high explainability. Like traditional spectral-based GNN, graph wavelet neural networks use diagonal matrix for graph filter. It is effective for node-level classification tasks (This task is giving the labels corresponding to some certain nodes in a given graph, thus predicting other nodes' labels), because different nodes' features don't need to be  mixed together. However, researchers need graph-level representation to finish graph classification tasks, which makes it difficult to apply wavelet neural networks to graph classification tasks.

Graph wavelet transform is becoming more frequently used in the design of GNNs. Xu et al.\cite{xu2019graph} designed the graph wavelet neural network(GWNN) using spectral graph theory for node classification tasks and obtained satisfactory results. They reported that using graph wavelet transform can circumvent the short-comings of previous spectral CNN methods, depending on the graph Fourier transform. Similarly to the graph Fourier transform, the wavelet base is designed in the following manner:
\begin{equation}
	\Psi_{s} = \mathbf{U_s}\mathbf{G_s}\mathbf{U_{s}^T},
\end{equation}
where $\mathbf{U_s}$ represents the Laplacian eigenvectors and $\mathbf{G_s}=diag(e^{\lambda_{1}s },...,e^{\lambda_{n}s })$ is the scaling matrix. Substituting the graph Fourier transform with wavelet transform, GWNN uses diagonal masks to generate the representation of each node. The structure of the $m$-th layer is defined as:
\begin{equation}
	\boldsymbol{X}_{[:, j]}^{m+1}=h\left(\psi_{s} \sum_{i=1}^{p} \boldsymbol{F}_{i, j}^{m} \psi_{s}^{-1} \boldsymbol{X}_{[:, i]}^{m}\right) \quad j=1, \cdots, q.
\end{equation}
Note that ${F}_{i, j}^{m}$ is a diagonal matrix, which is effective for node-level classification tasks, as the features of different nodes cannot be mixed. GWNN is highly competitive at node-level tasks. However, GWNNs don't have a mechanism to handle graphs of different sizes, which is crucial for graph classification tasks. Besides, GWNNs lack a standard pooling mechanism to aggregate node-level features into a fixed-size graph-level representation, which is necessary for classifying graphs of varying sizes. 

As the application in graph multi-modal learning, Behmanesh et al.\cite{behmanesh2022geometric} proposed a graph wavelet convolution network (GWCN) for multi-modal learning. GWCN generates single-modal representations by applying the multi-scale graph wavelet transform and learning permutations that encode correlations among various modalities. GWCN have the best performance on node classification tasks. 

Wavelet-based methods are a powerful tool for capturing multi-scale graph representations. However, currently, few methods use graph wavelet transform for cross-scale graph classification. 
\section{GSpect}
\subsection{Problem Description}\label{def}

\begin{figure}[htb]
	\includegraphics[width=0.5\textwidth]{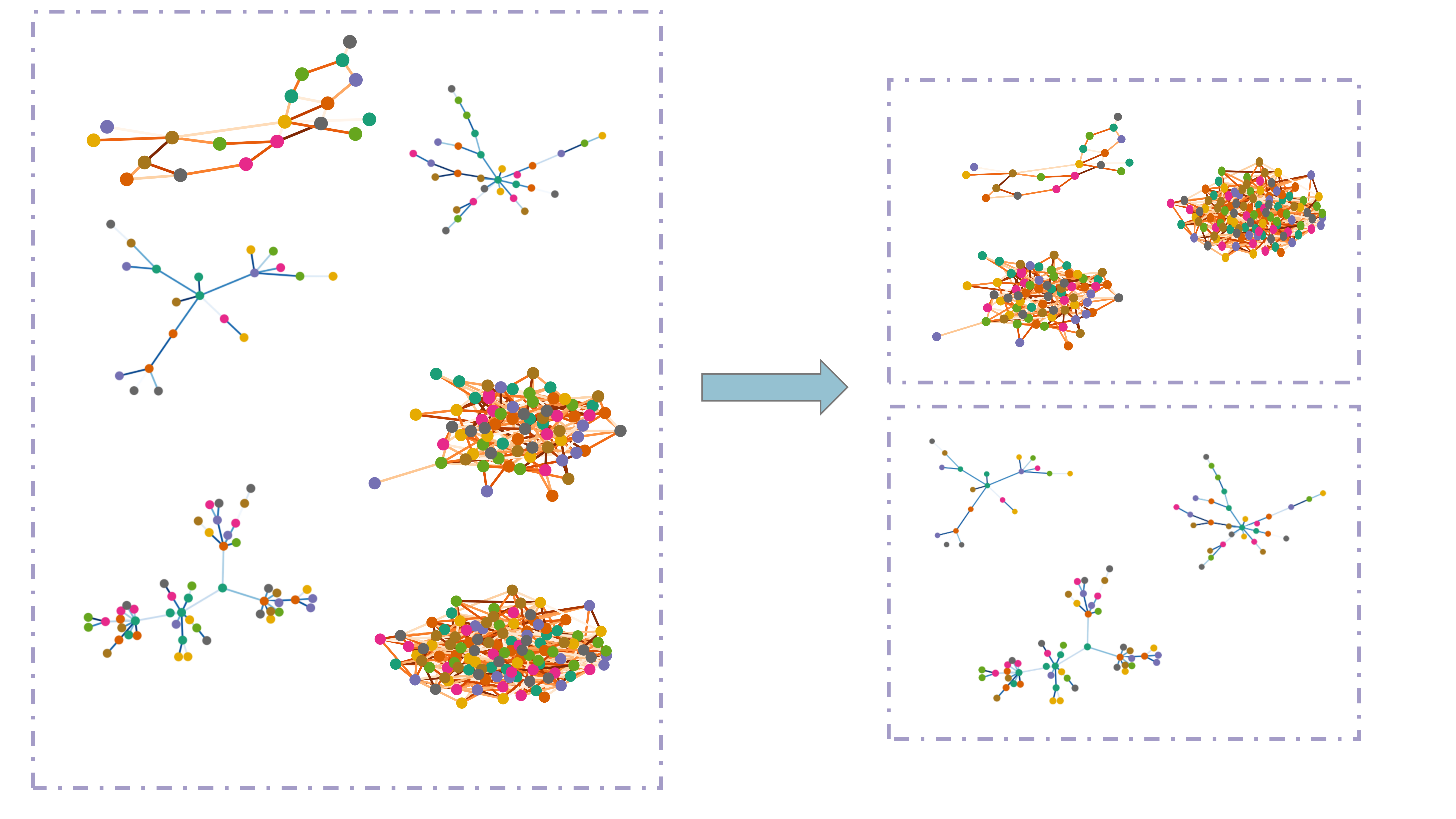}
	\caption{An example of cross-scale graph classification } \label{fig1}
\end{figure}

Let $G = \{V,E\}$ represent a graph, with $V$ and $E$ being the set of nodes and edges, respectively. $A\in \left \{ 0,1 \right \} ^{n\times n}$ represents the adjacency matrix and $X\in\mathbb{R} ^{n\times l}$ represent the node attribute matrix. $l$ represents the length of the attribute vector. There is a set of labeled graphs $(\{G\},\{y\}) $, where $y_i \in \mathbb{Z}$ represents the label of $G_i$, and max[size($ \{G\} $)]/min[size($ \{G\} $)] $\geq 10^3 $. The target of the cross-scale graph classification task is to learn a mapping $f:G\to y $. Compared with other machine learning methods applied in computer vision and natural language processing, we need to convert graphs with different topologies into vector $v\in \mathbb{R}^q$, where $q\le  min(n)$. Then, the mature approach of machine learning methods can be used. Fig. \ref {fig1} depicts an example of cross-scale graph classification. 
\subsection{Model Framework}
\begin{figure*}[htb]
	\includegraphics[width=\textwidth]{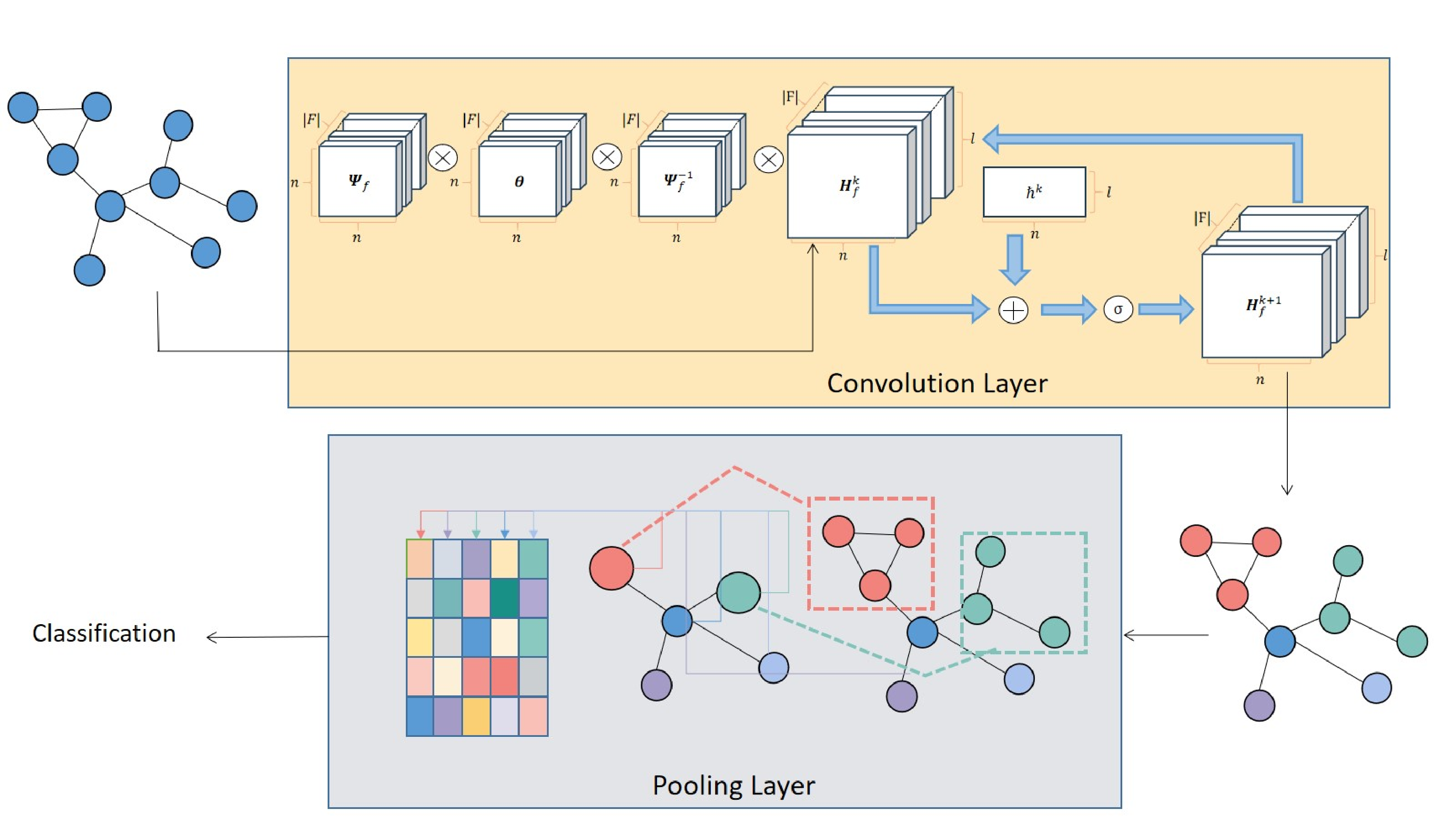}
	\caption{The model structure of GSpect. GSpect consists of four phases. The first phase consists of the convolution layers. Each layer has F multi-scale graph wavelet convolution. The second phase is a pooling layer. This layer aggregates the nodes with similar representations in the spectrum and yields a low-order graph. The third phase is a full-connect layer for classification. the colored matrix indicate the feature vector of each node. In the second phase, nodes with similar features (depicted as the same color in the diagram) are aggregated into a single node.  } \label{fig2}
\end{figure*}

In this section, we introduce the framework of our model GSpect. GSpect consists of four parts(Fig. \ref {fig2}). The first part is the convolution layer. We use graph wavelet transform for the convolution layer to generate the graph-level representation. In the second part, we design the spectral-pooling layer to filter the useless information and obtain the low-order representation for classification. The spectral-pooling layer aggregates the nodes with similar representations in the spectrum and obtain a low-order graph. The third part is a fully connected layer for classification. Because the convolution and pooling process need to be repeated many times, we use simple GCN in convolution after pooling. Finally we design an optimising function to optimise the model. Furthermore, we proved the stability of the model (see Appendix \ref{sta}).

\subsection{Graph Wavelet Convolution Layer}
As the first step of GSpect, we design a convolution layer to generate the graph presentations. For traditional convolution methods, cross-scale graphs has big difference in size, which leads to the difficulty of getting graph presentations. In this section, we propose the graph wavelet convolution layer (GWC). Taking advantage of the fact that the wavelet function can capture multi-scale messages, we use the wavelet transform to project the graph into the wavelet domain and use a learnable filter to aggregate messages from every entry and obtain the graph representation. 

In earlier research, wavelet bases are defined in the following manner: $\Psi_{f}=\left[\psi_{f, 1}, \ldots, \psi_{f, N}\right] $, where $\psi_{f, i}$ represent the the transform matrix at node $i$ and scale $f$. Different studies have various definitions of $\psi_{f, i}$. A few traditional functions of wavelet bases need to compute the eigenvalues of the graph, which leads to a large amount of computation. To escape this, we use the definition of \cite{hammond2011wavelets} to approximate the wavelet bases, which is defined in the following manner:

\begin{align}
	\Psi_{f}=\frac{1}{2} c_{0, f}+\sum_{i=1}^{M} c_{i, f} \mathbf{T}_{i}(\tilde{\mathbf{L}}),
	\\
	c_{i, f}=2 e^{-f} J_{i}(-f), \ \ \ \ \ \
	\label{psi}
\end{align}
where $\tilde{\mathbf{L}}$ is the Chebyshev polynomial\cite{debnath1991structure} of order $i$ which is used to approximate $\Psi_{f}$, $M$ is the number of Chebyshev polynomials and $J_{i}(-f)$ is the Bessel function of the first category\cite{arfken2011mathematical} and $f$ is the wavelet scale. 

sAccording to prior research\cite{behmanesh2022geometric}, we use the wavelet base $\Psi_{f}^{-1}$ to project the graph's embedding matrix to the wavelet domain. 

Since the formula Equation \ref{psi} is in an approximate form, the inverse of the matrix may not exist. Therefore, this article uses the pseudoinverse of the matrix instead. We first perform singular value decomposition on $\Psi_{f}$, that is:

\begin{equation}
	\Psi_{f} = V \Sigma U^T.
\end{equation}
Where $V$ and $U$ are left and right singular vector matrix. 
Then the inverse $\Psi_{f}^{-1}$ is defined as follows.

\begin{equation}
	\Psi_{f}^{-1} = V \Sigma^{-1} U^T  
\end{equation}

Then, in the wavelet domain we use a learnable filter to aggregate messages from every entry. Thereafter, we use  $\Psi_{f}$ to convert the representation back. Finally, we use the bias and activation functions to formalise the convolution layer. The one-scale-channel convolution layer is defined in the following manner:
\begin{equation}
	H_{n\times l,f}^{k+1}  =  \sigma (\Psi _{n\times n,f}  \Theta_{n\times n}  \Psi _{n\times n,f}^{-1} H_{n\times l,f}^{k}+\hbar_{n\times l} ),
\end{equation}
where $n$ represents the node number, $f$ represents the wavelet scale, and $k$ represents the k-th layer. $\Theta$ and $\hbar$ are learnable parameters. 
There are many scales which are responsible for aggregating messages on their own scale. By averaging the messages of $F$ scales, the total convolution layer is defined in the following manner:
\begin{equation}
	H_{n\times l}^{k}  =  \frac{1}{F} \sum_{f=1}^{F}  H_{n\times l,f}^{k} .
\end{equation}
We use average graph representation by averaging the graph representations of all scales, which synthesise the graph structure messages on different scales.

\subsection{Spectral-pooling Layer} 
Since the graphs have different sizes even after convolution, they cannot be directly classified. To solve this problem, we design a pooling layer to process the graph presentation and generate graphs in the same size for classification. 

Motivated by the research\cite{ma2019graph}\cite{ying2018hierarchical}, we continue to use the concept assignment matrix:

\begin{equation}
	S^{k} = GNN_{k,pooling}(A^{k}, X^{k}) ,
\end{equation}
which implies learning a project matrix that projects the adjacency matrix to a low-order adjacency matrix. In essence, it converges a group of nodes to a single node. Rather than using a normal GNN structure to learn $S^{k}$ directly, we propose a new method in this article. We use Fourier transform to convert the adjacency matrix $A$ and graph embedding $X$ into a frequency domain and use a spectral filter to filter out useless information and reduce the size of matrix through spectral convolution. The assign matrix is defined in the following manner:
\begin{equation}
	S^{k}_{(n-m)\times n} =\xi^{k} _{(n-m)\times(n-m)  } \theta _{(n-m)\times n}^{k} \xi^{-1,k} _{n\times n  },
\end{equation}
where $\xi(u, v) = \sum_{x} \sum_{y} f(x, y) e^{-j 2 \pi \left(\frac{u x}{M} + \frac{v y}{N} \right)}$  is the Fourier transform matrix. The function $f(x, y)$ can be any arbitrary function. $n$ and $m$ is the node number before pooling and after. $\theta _{(n-m)\times n}$ is the learnable parameter.
Thus, the total equation of adjacency matrix $A$ and graph embedding $X$ is:
\begin{align}
	X_{(n-m)\times l}^{k+1} = S_{(n-m)\times n} X_{n\times l}^{k} ,\ \ \ \ \ \ \ \ \
	\\
	A_{(n-m)\times (n-m)}^{k+1} = S_{(n-m)\times n}^{k}   A_{n\times n}^{k}(S_{(n-m)\times n}^{k})^{T}. 
\end{align}
%The pooling process in wavelet domain is the mixture of different values in spectrum. Fig. \ref{fig5} shows the pooling process in spectrum. The figure above is the pooling process in physical domain. The figure below is the pooling process in wavelet domain. The boxes with different colors refer to the graphs' figure of spectrum getting from Fourier transform. The different colors of the box refer to the different values in frequency band. During the pooling process, the values of different bands generate fusion
%\begin{figure}[htb]
%	\includegraphics[width=\textwidth]{./examples/total.eps}
%	\caption{} \label{fig5}
%\end{figure}
\subsection{The Optimization Method} 
The parameters in the model need to be optimized. In this section, we introduce the optimization function of GSpect. According to existing research\cite{liu2022graph}, it is difficult to optimize the model using gradient descent only during the graph classification task. To solve this question, we use the weighted optimization function. We will introduce the optimization functions separately.

Cross entropy is an important concept in information theory. Its value represents the difference between two probability distributions. The approximate of the target probability distribution can be obtained by minimizing cross entropy. First, we use the cross entropy function as a part of our optimization function, which is defined in the following manner:
\begin{equation}
	L_{\varepsilon}(p,q) = \frac{1}{c} \sum_{i=1}^{c} p_ilog(q_i),
\end{equation}
where $p_i$ and $q_i$ are true and predicted labels, and $c$ is the class number.

The assign matrix should meet one condition: the nodes having strong links have higher probability of aggregating to a new node\cite{ying2018hierarchical}. Thus the second part of the optimization function is expressed in the following manner:
\begin{equation}
	L_{p} = \left \| A^k - S^k(S^k)^T  \right \| _F,
\end{equation}
where $\left \| . \right \| _F$ represents the Frobenius norm. This equation implies to let $A^k$ and $ S^k(S^k)^T$ be as close as possible. Specifically, for $ A^{k} $, when $ k=0 $, $ A^{0} $  is the graph's adjacency matrix. When $ k \neq 0 $, $ A^{k} $is the processed adjacency matrix in the $ k $-th layer. $ A_{i j}^{(k)} $ represents the link between node $ i $ and node $ j $ in the $ k $-th layer.

For $ S^{(k)} S^{(k)^{T}} $, $ S^{(k)} \in \mathbb{R}^{n_{k} \times n_{k+1}}\left(n_{k}>n_{k+1}\right) $ is the probability matrix, $ S_{i r}^{(k)} $ represents the probability of node $ i $ in the $ k $-th layer, thereby mapping to node $ j $ from cluster $ r $ in $ (l+1) $-th layer. 
%$ \left(S^{(k)} S^{(k)^{T}}\right)_{i j}=\Sigma_{r} S_{i r}^{(k)} S_{r j}^{(k)^{T}} $  represent i-th row in S^{(l)}  的第  i  行和  S^{(l)^{T}}  的第  j  列对应相乘再相加，也可表示为  S^{(l)}  的第  i  行和第  j  行对应相乘再相加，即  i  节点和  j  节点映射 到下一层同一个cluster的概率对应相乘再相加。
When the probability of two nodes mapping to one cluster increases, the value of $ S^{(l)} S^{(l)^{T}} $ becomes larger. 
Minimizing $ L_{p} $ and retaining the correct assignment matrix $ S^{(l)} $ can let the pair of nodes that has a stronger link easily map to one cluster. 

Thus, the total optimisation function is expressed in the following manner:
\begin{equation}
	L_t = (1-\beta)L_{\varepsilon} + \beta L_{p} \label{ballence},
\end{equation}
where $\beta$ is the equilibrium coefficient.
\section{EXPERIMENT}
\begin{table*}[!htb]
	\centering
	
	\caption{Basic Statistics of the Open Data Sets and MSG }\label{tab3}
	
	\begin{tabular}{ c c c c c c c}
		
		%		\hline \text { Name of Data Set } & \text { Number of Graphs} & \text {Number of Max Nodes } & \text {Number of Average Nodes } \\
		%		
		%		$ 210 $ & $ 1000 $ & $ 228.67 $ \\
		%		\hline 
		%		\text { S.D. of Node Distribution } & \text { Average Degree } & \text { Class Number}\\
		%		$ 326.28 $ & $ 14.04 $ & $ 6 $\\
		\hline 
		\text {Name} & \text {Avg Graph Size} &\text {Avg Degree} & \text {Avg Edges Number} & \text { Min~Max Graph Size} &  \text {S.D. of Node Distribution}& \text{Avg Network Diameter}\\ \hline
		PTC & $ 25.56 $ & $ 1.99 $ & $ 25.25 $ & $ [2,109] $ & $ 16.25 $ & $ 8.78 $\\ 
		MUTAG & $ 17.93 $ & $ 2.19 $ & $ 19.79 $ & $ [10,28]  $ & $ 4.58 $ & $ 8.22 $\\ 
		PROTEINS & $ 39.05 $ & $ 3.73 $ & $ 57.72 $ & $[4,620]$ & $45.76$ & $ 10.75 $\\ 
		D\&D & $268.70 $ & $ 4.98 $ & $ 173.98 $ & $ [30,903] $ & $ 161.33 $ & $ 12.14 $\\ 
		IMDB-B & $ 19.77 $ & $ 8.89 $ & $ 95.38 $ & $ [12,136] $ & $ 10.06 $ & $ 1.86 $\\ 
		\hline 
		MSG class-1 & $ 49.43 $ & $ 3.66 $ & $ 88.20 $ & $ [5,150] $ & $ 42.22 $ & $ 14.33 $\\
		MSG class-2 & $ 33.67 $ & $ 3.64 $ & $ 61.93 $ & $ [4,100] $ & $ 27.02 $ & $ 10.80 $\\		
		MSG class-3 & $ 379.96 $ & $ 66.10 $ & $ 24238.56 $ & $ [4,1000] $ & $ 369.98 $ & $ 3.48 $\\
		MSG class-4 & $ 332.00 $ & $ 4.00 $ & $ 664.00 $ & $ [10,1000] $ & $ 377.22 $ & $ 25.97 $\\  
		MSG class-5 & $ 21.17 $ & $ 8.73 $ & $ 115.83 $ & $ [12,65] $ & $ 11.80 $ & $ 1.93 $\\
		MSG class-6 & $ 524.65 $ & $ 2.00 $ & $ 524.85 $ & $ [49,1000] $ & $ 288.22 $ & $ 13.85 $\\	
		\hline
		
	\end{tabular}
\end{table*}
In this section, we test the model's effectiveness on graph classification tasks.  We aim to answer the following questions: 

$Q1$ \ How does our model compared to other advanced models in open data sets?

$Q2$ \ To what extent does our model improve the performance of a baseline GNN?

$Q3$ \ Is GSpect sensitive to changes in hyperparameters?

The code and other materials are available at  \url{https://github.com/XiaoyuZhang001/GSpect}.
\subsection{Experiment Settings} 
\subsubsection{Data Sets} 
We use the following five open data sets to verify the effectiveness of the model:

D\&D \cite{dobson2003distinguishing} (Biological macromolecules). D\&D is a protein data set. It extracted 1178 high-resolution proteins from a non-redundant subset of the protein database using simple features, such as secondary structure content, amino acid propensity, surface properties, and ligands. The nodes are amino acids, and if the distance between the two nodes is less than six angstroms, an edge is used to represent this relationship. Nodes in DD data set are unlabeled and nodes only have features. The criterion for classification is whether a protein is an enzyme. 

PTC\cite{toivonen2003statistical} (Small molecules). PTC is a collection of 344 compounds that report carcinogenicity to rats. Researchers need to classify these compounds to the criterion of carcinogenicity. Nodes represent atoms and edges between nodes represent bonds between corresponding atoms. Each node has 19 node labels. 

PROTEINS\cite{borgwardt2005protein} (Biological macromolecules). PROTEINS is another network of proteins. The task is to determine whether such molecules are enzymes. The nodes are amino acids.

IMDB-B\cite{yanardag2015deep} (Social network). IMDB-B is a movie collaboration data set consisting of a self-network of 1,000 actors who play movie roles in IMDB. In each network, the nodes represent the actors/actresses. Researchers use an edge to link them if they act in the same movie. The criterion for classification is the type of movies. These networks are collected from the action movies and romantic movies. 

MUTAG\cite{debnath1991structure} (Small molecules). MUTAG is a data set of nitroaromatic compounds designed to predict their mutagenicity against salmonella typhimurium. The graphs are used to represent compounds, where nodes represent atoms and are labeled by atomic type (represented by single encoding), while edges between nodes represent bonds between corresponding atoms. It includes 188 compound samples and 7 discrete node labels.
%\begin{table*}[!htb]
%	\centering
%	
%	\caption{Basic Statistics of the Open Data Sets and MSG }\label{tab3}
%	
%	\begin{tabular}{ c c c c c c c}
%		
%%		\hline \text { Name of Data Set } & \text { Number of Graphs} & \text {Number of Max Nodes } & \text {Number of Average Nodes } \\
%%		
%%		$ 210 $ & $ 1000 $ & $ 228.67 $ \\
%%		\hline 
%%		\text { S.D. of Node Distribution } & \text { Average Degree } & \text { Class Number}\\
%%		$ 326.28 $ & $ 14.04 $ & $ 6 $\\
%		\hline \text {Name} & \text {Avg Graph Size} &\text {Avg Degree} & \text {Avg Edges Number} & \text { Min~Max Graph Size} &  \text {S.D. of Node Distribution}& \text{Avg Network Diameter}\\ \hline
%		PTC & $ 25.56 $ & $ 1.99 $ & $ 25.25 $ & $ [2,109] $ & $ 16.25 $ & $ 8.78 $\\ 
%		MUTAG & $ 17.93 $ & $ 2.19 $ & $ 19.79 $ & $ [10,28]  $ & $ 4.58 $ & $ 8.22 $\\ 
%		PROTEINS & $ 39.05 $ & $ 3.73 $ & $ 57.72 $ & $[4,620]$ & $45.76$ & $ 10.75 $\\ 
%		D\&D & $268.70 $ & $ 4.98 $ & $ 173.98 $ & $ [30,903] $ & $ 161.33 $ & $ 12.14 $\\ 
%		IMDB-B & $ 19.77 $ & $ 8.89 $ & $ 95.38 $ & $ [12,136] $ & $ 10.06 $ & $ 1.86 $\\ 
%		\hline 
%		MSG & $ 228.67 $ & $ 11.78 $ & $ 3210.67 $ & $ [4,1000] $ & $ 326.28 $ & $ 14.42 $\\ \hline 
%		
%		
%	\end{tabular}
%\end{table*}

In this article, we collect a number of empirical networks---including the set of protein structure data, macromolecular compound structure data, and social networks data---in combination with the three typical modeled networks of BA\cite{barabasi1999emergence}, WS\cite{watts1998collective} and ER\cite{erdHos1960evolution} to create a synthesis cross-scale graph classification benchmark data set MSG. Table \ref{tab3} presents the basic statistical properties of open data sets and MSG. The large standard deviation of the node distribution reflects the large difference in the size of the graphs, which reflects the goal of cross-scale graph classification. A visual comparison of structures with varying sizes in different classes are depicted in Fig. \ref {fig3}.

\begin{figure}[htb]
	\includegraphics[width=0.5\textwidth]{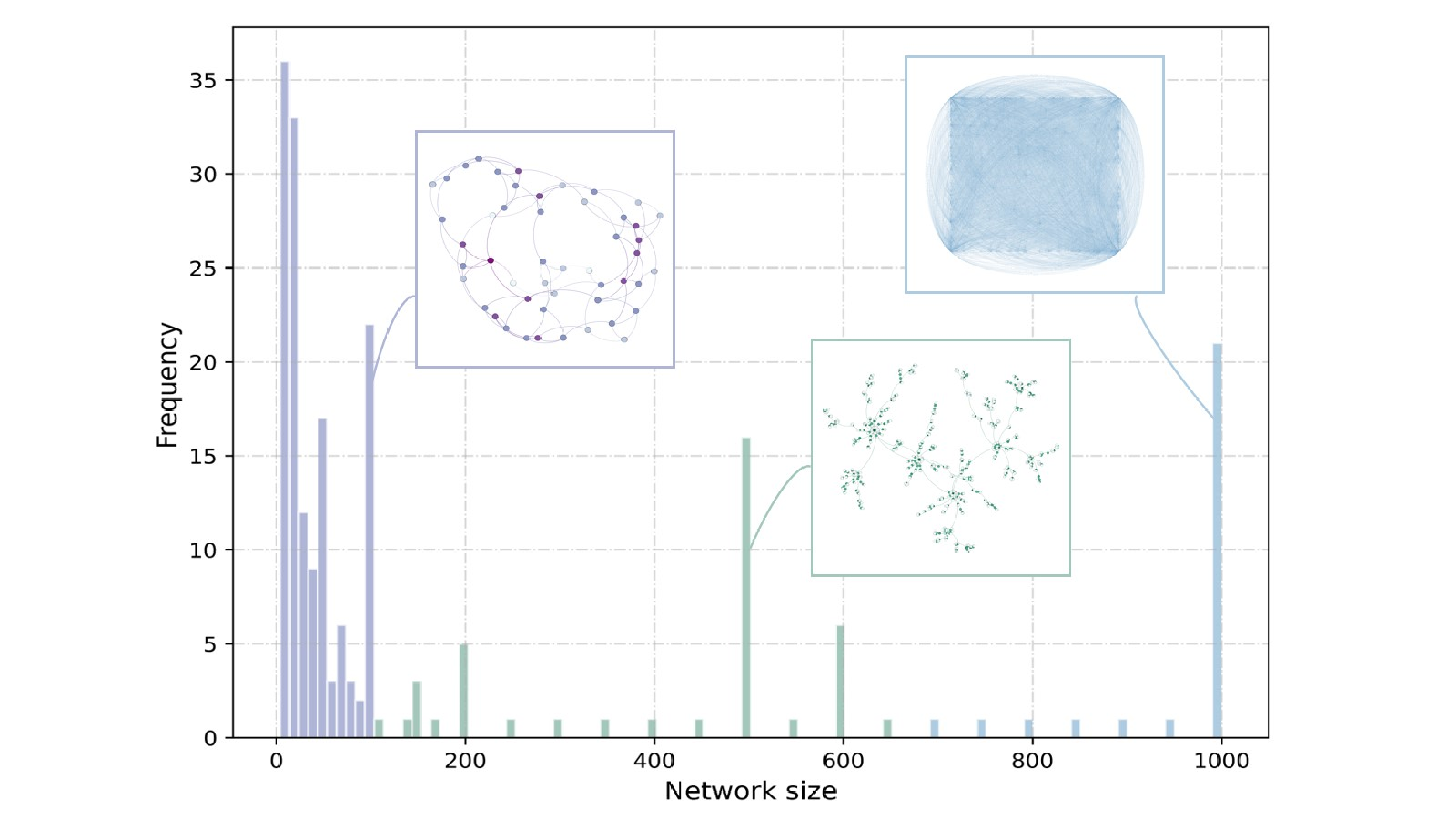}
	\caption{The distribution of of the graph size of the MSG data set} \label{figdis}
\end{figure}

\begin{figure*}[htp]
	%	\flushleft
	\centering  %图片全局居中
	\vspace{-0.35cm} %设置与上面正文的距离
	%	\subfigtopskip=2pt %设置子图与上面正文或别的内容的距离
	%	\subfigbottomskip=2pt %设置第二行子图与第一行子图的距离，即下面的头与上面的脚的距离
	%	\subfigcapskip=-5pt %设置子图与子标题之间的距离
	
	\subfloat[class 1-1]{
		\begin{minipage}[t]{0.25\linewidth}
			\centering
			\includegraphics[width=3.5cm,height=3.5cm]{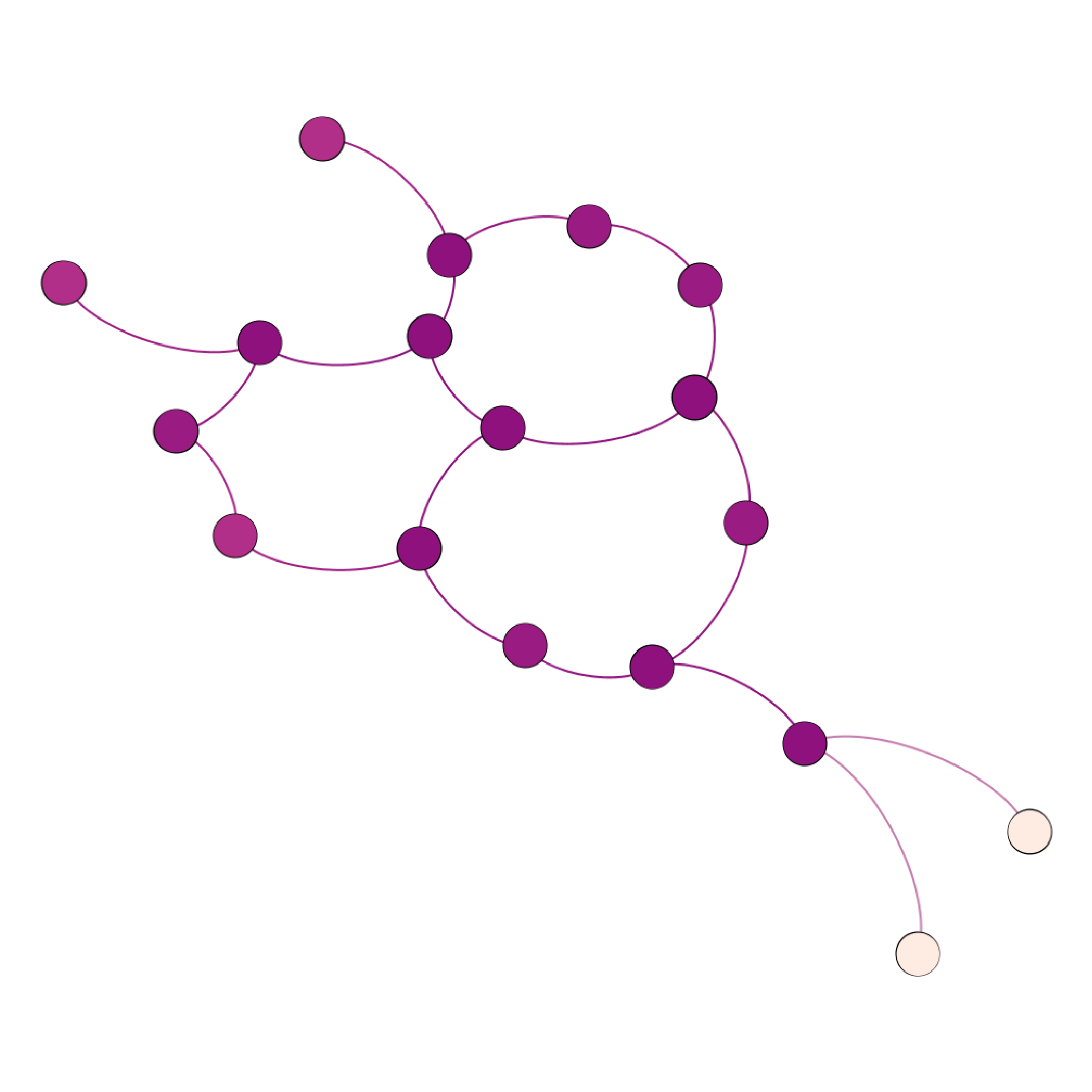}
			
		\end{minipage}
	}
	\subfloat[class 1-2]{
		\begin{minipage}[t]{0.25\linewidth}
			\centering
			\includegraphics[width=4cm,height=4cm]{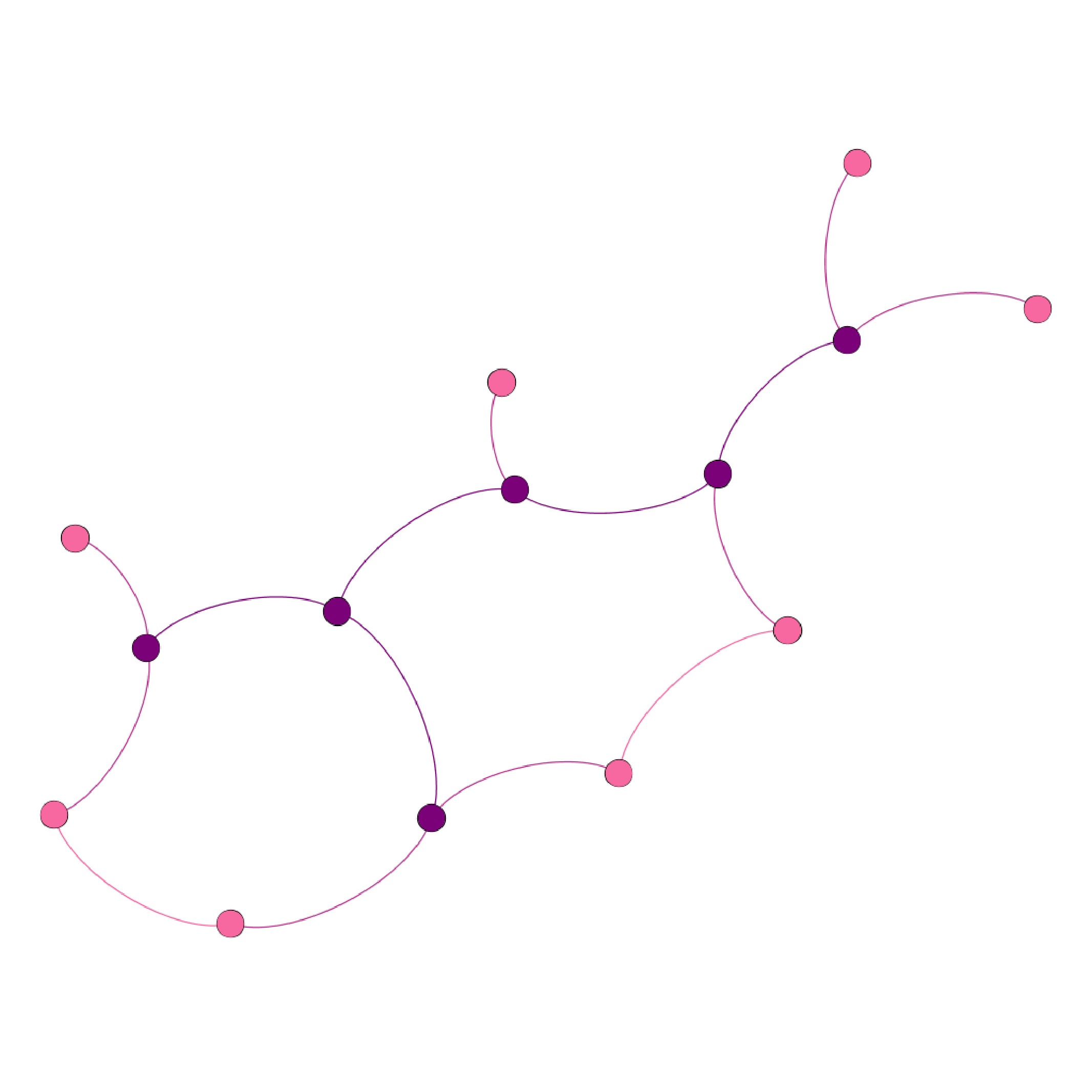}
			
		\end{minipage}
	}
	\subfloat[class 2-1]{
		\begin{minipage}[t]{0.25\linewidth}
			\centering
			\includegraphics[width=4cm,height=4cm]{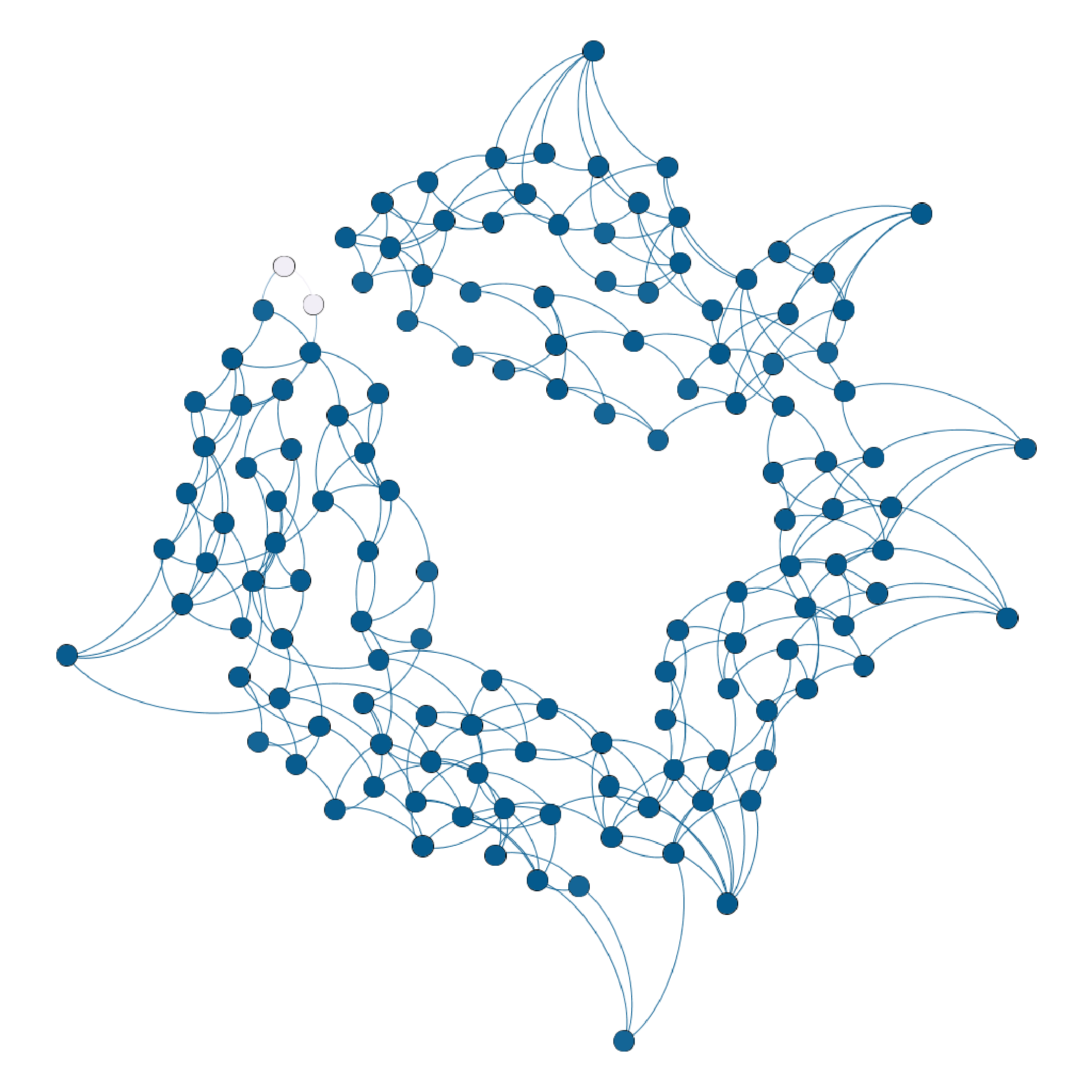}
			
		\end{minipage}
	}
	\subfloat[class 2-2]{
		\begin{minipage}[t]{0.25\linewidth}
			\centering
			\includegraphics[width=4cm,height=4cm]{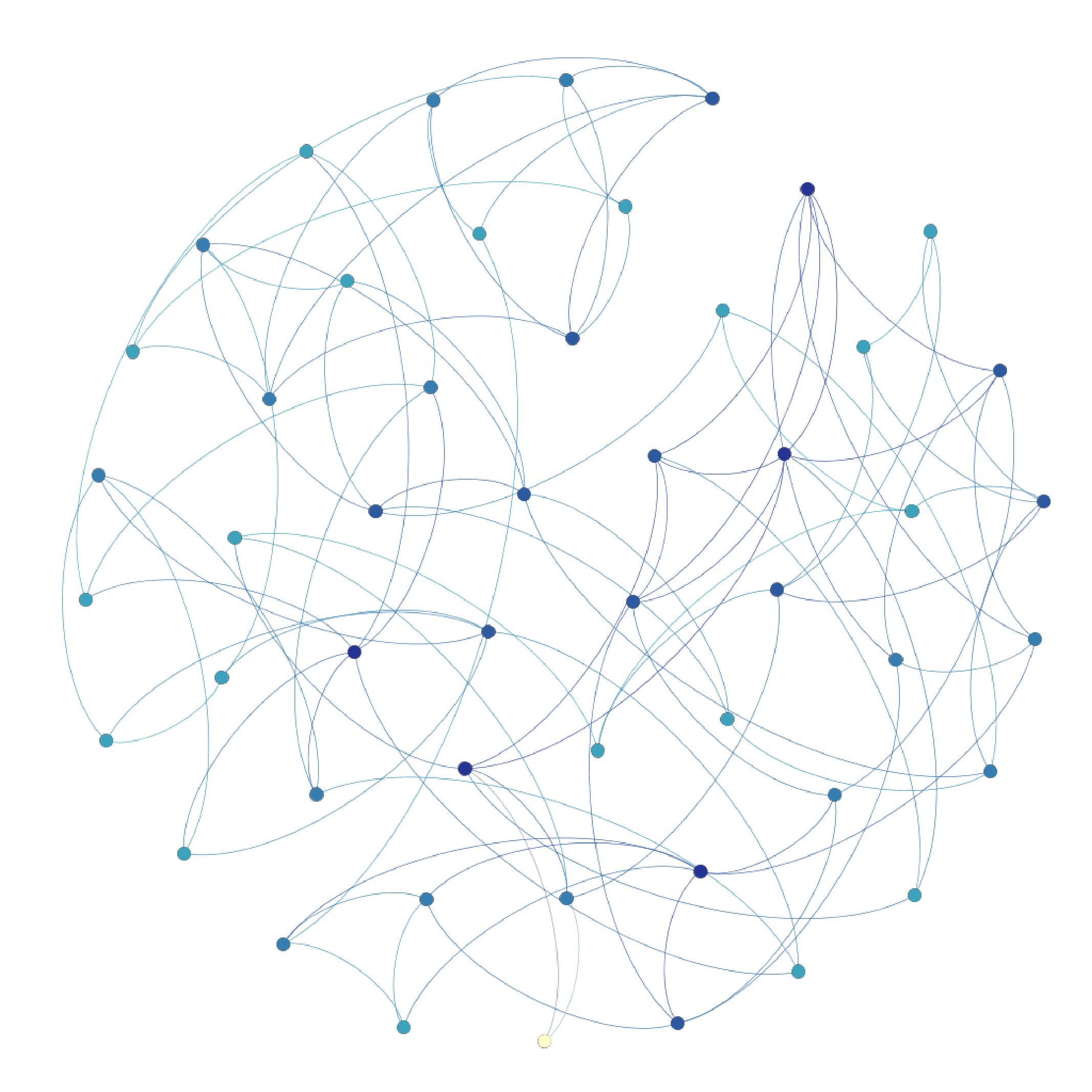}
			
		\end{minipage}
	}
	
	\subfloat[class 3-1]{
		\begin{minipage}[t]{0.25\linewidth}
			\centering
			\includegraphics[width=4cm,height=4cm]{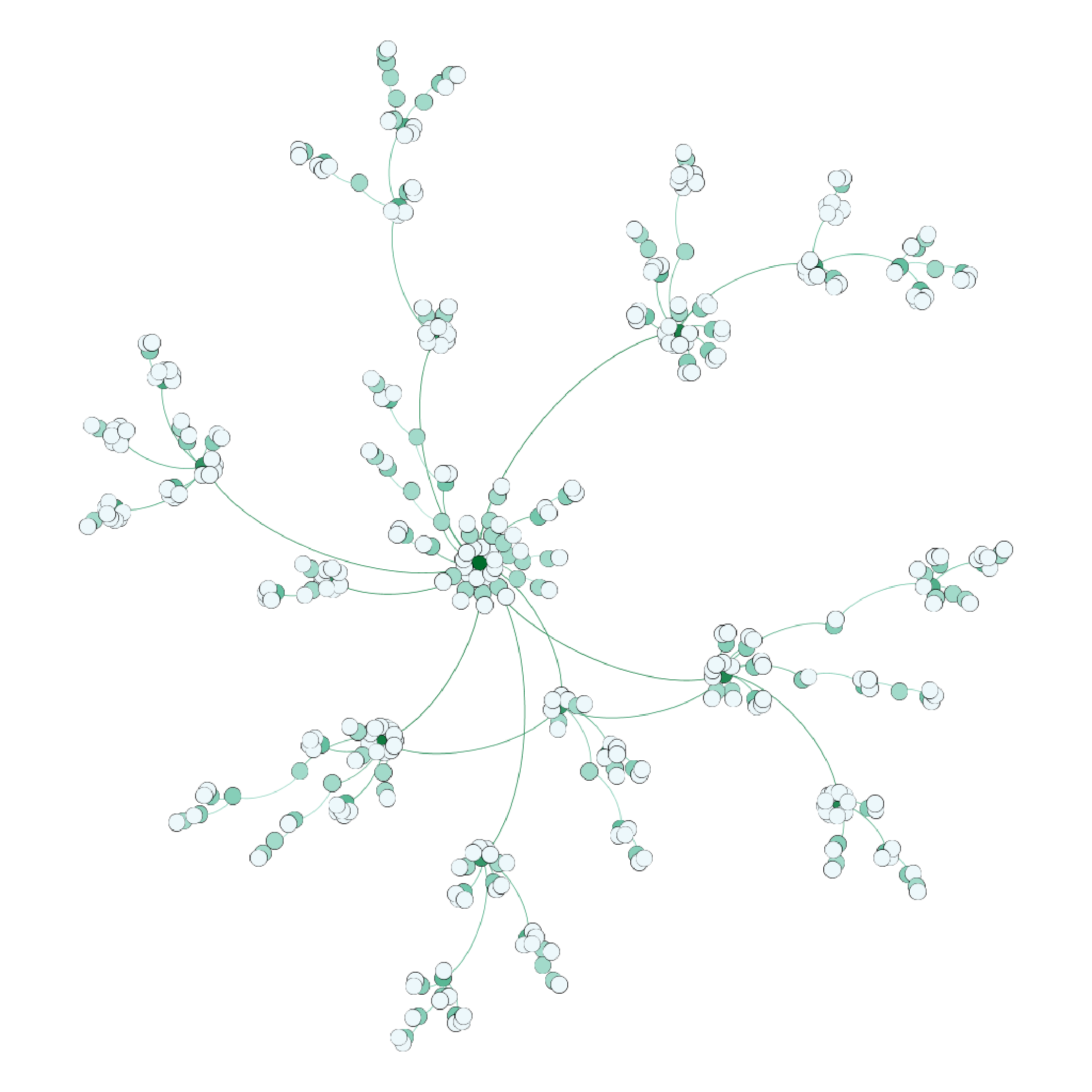}
			
		\end{minipage}
	}
	\subfloat[class 3-2]{
		\begin{minipage}[t]{0.25\linewidth}
			\centering
			\includegraphics[width=4cm,height=4cm]{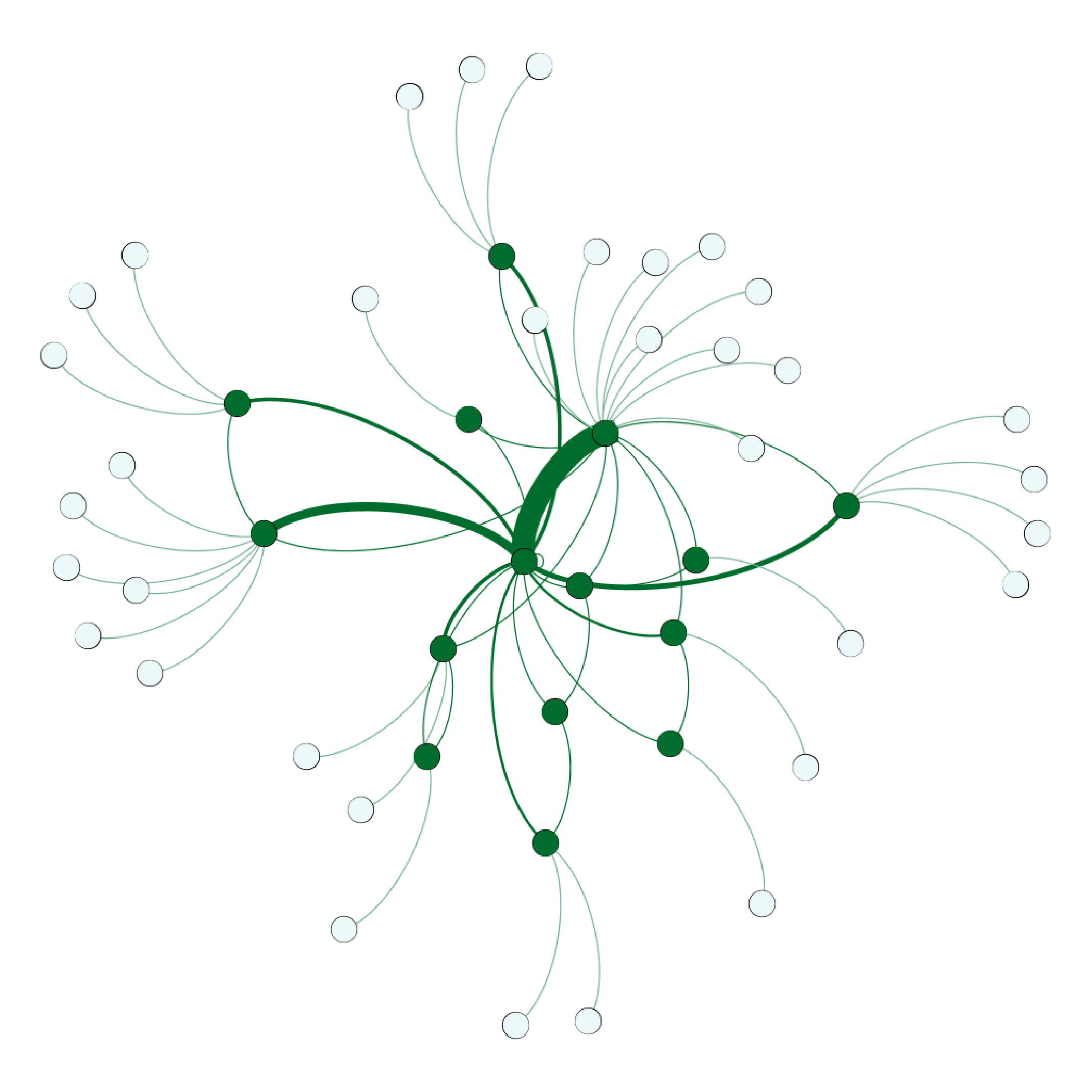}
			
		\end{minipage}
	}
	\subfloat[class 4-1]{
		\begin{minipage}[t]{0.25\linewidth}
			\centering
			\includegraphics[width=4cm,height=4cm]{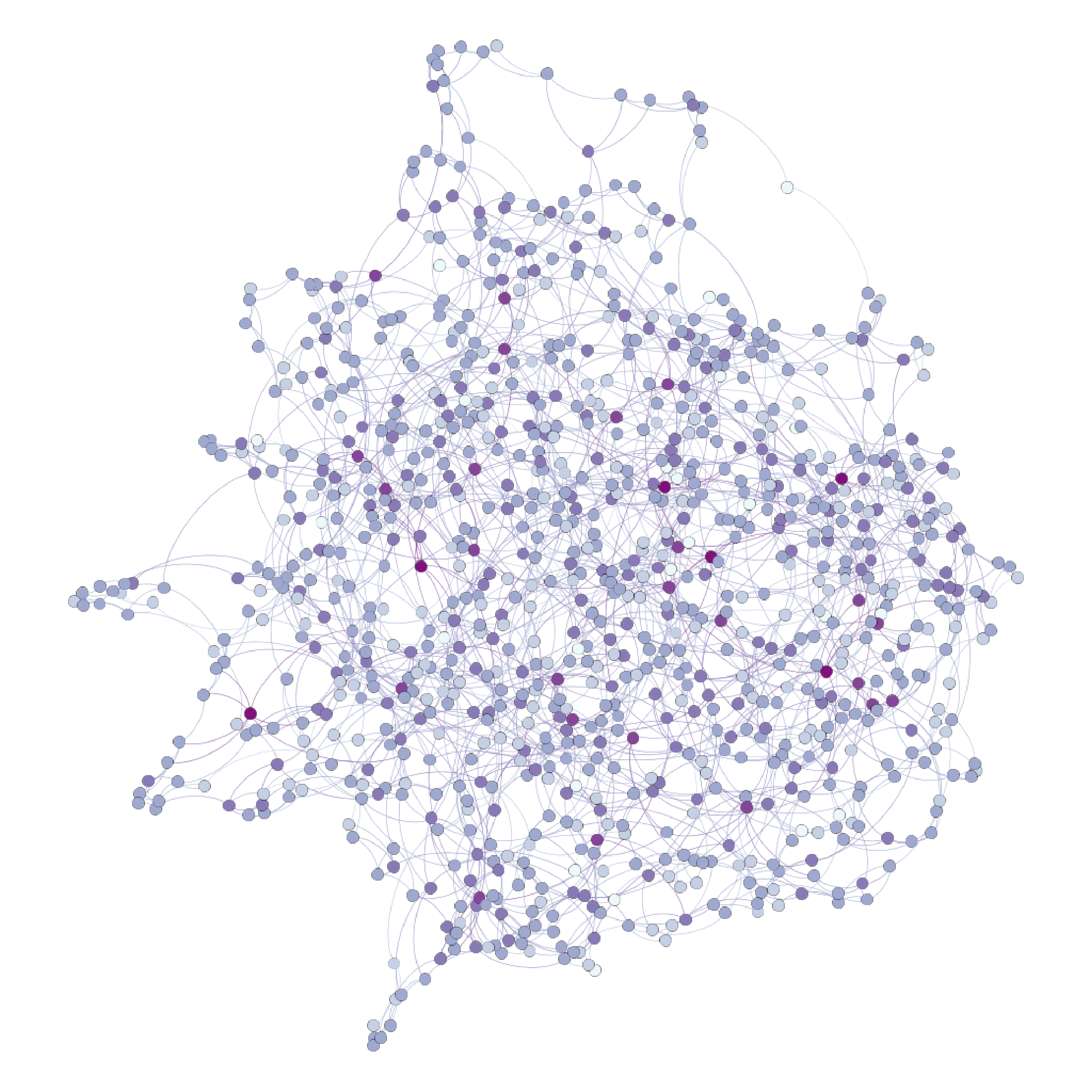}
			
		\end{minipage}
	}
	\subfloat[class 4-2]{
		\begin{minipage}[t]{0.25\linewidth}
			\centering
			\includegraphics[width=4cm,height=4cm]{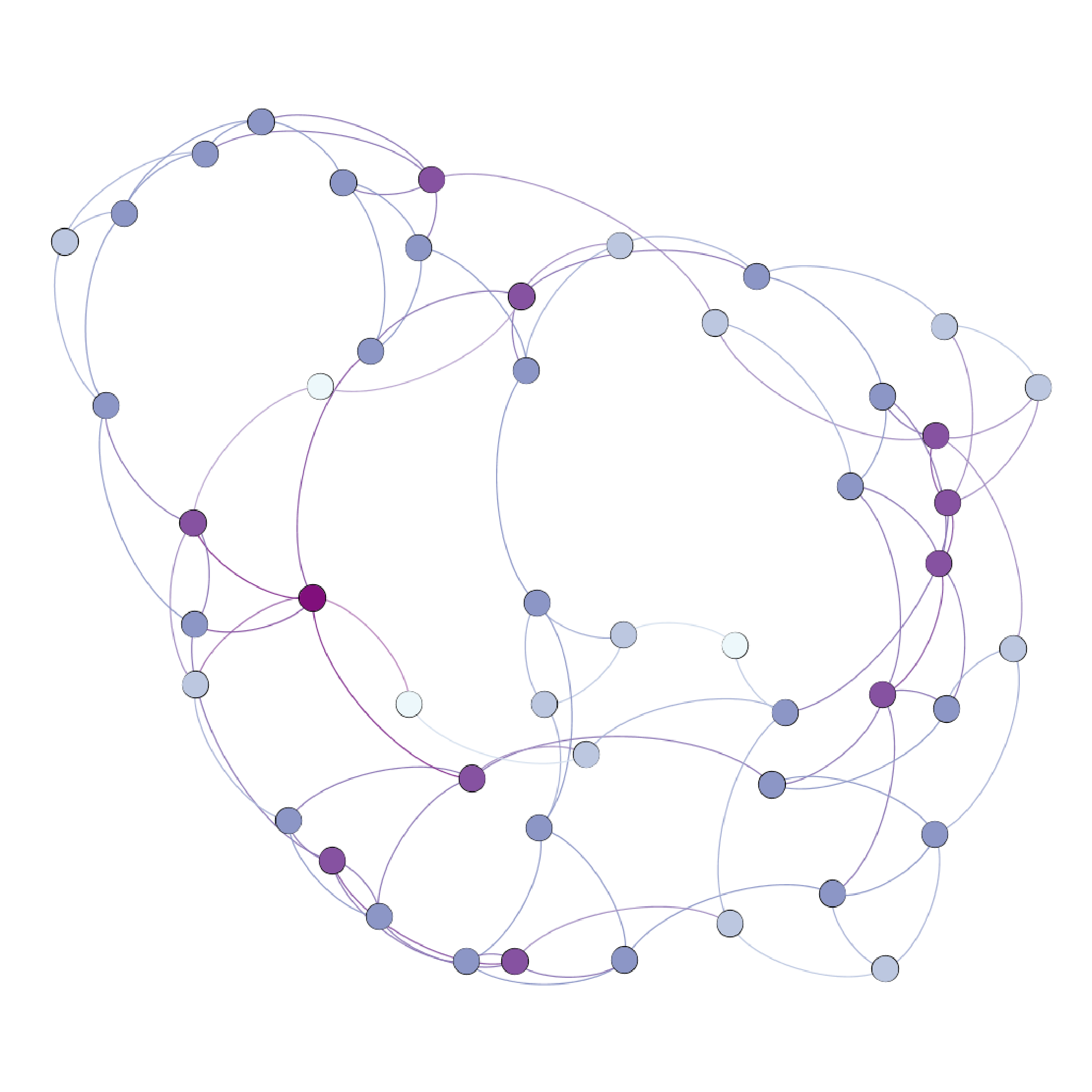}
			
		\end{minipage}
	}

	\subfloat[class 5-1]{
		\begin{minipage}[t]{0.25\linewidth}
			\centering
			\includegraphics[width=4cm,height=4cm]{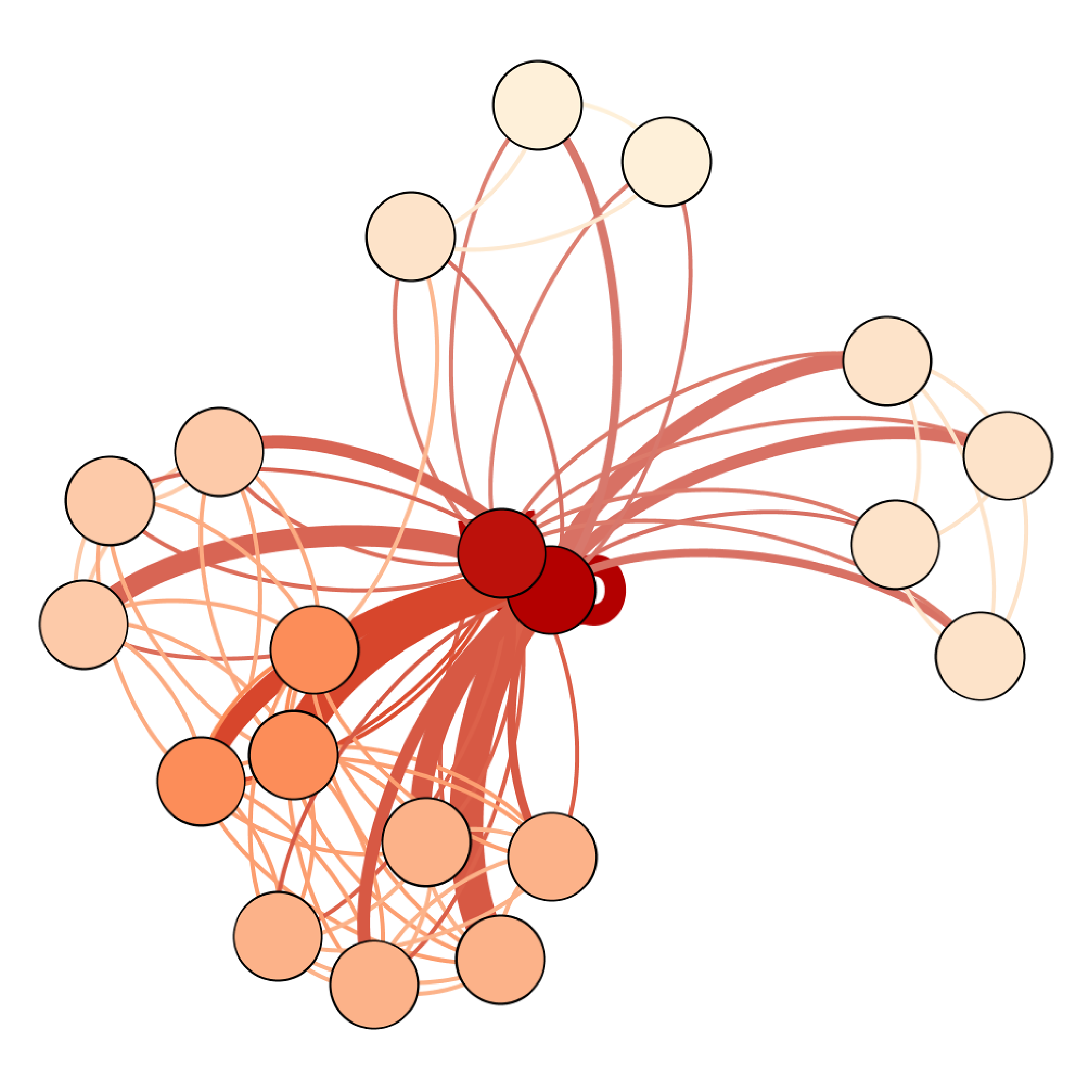}
			
	\end{minipage}	}
	\subfloat[class 5-2]{
		\begin{minipage}[t]{0.25\linewidth}
			\centering
			\includegraphics[width=4cm,height=4cm]{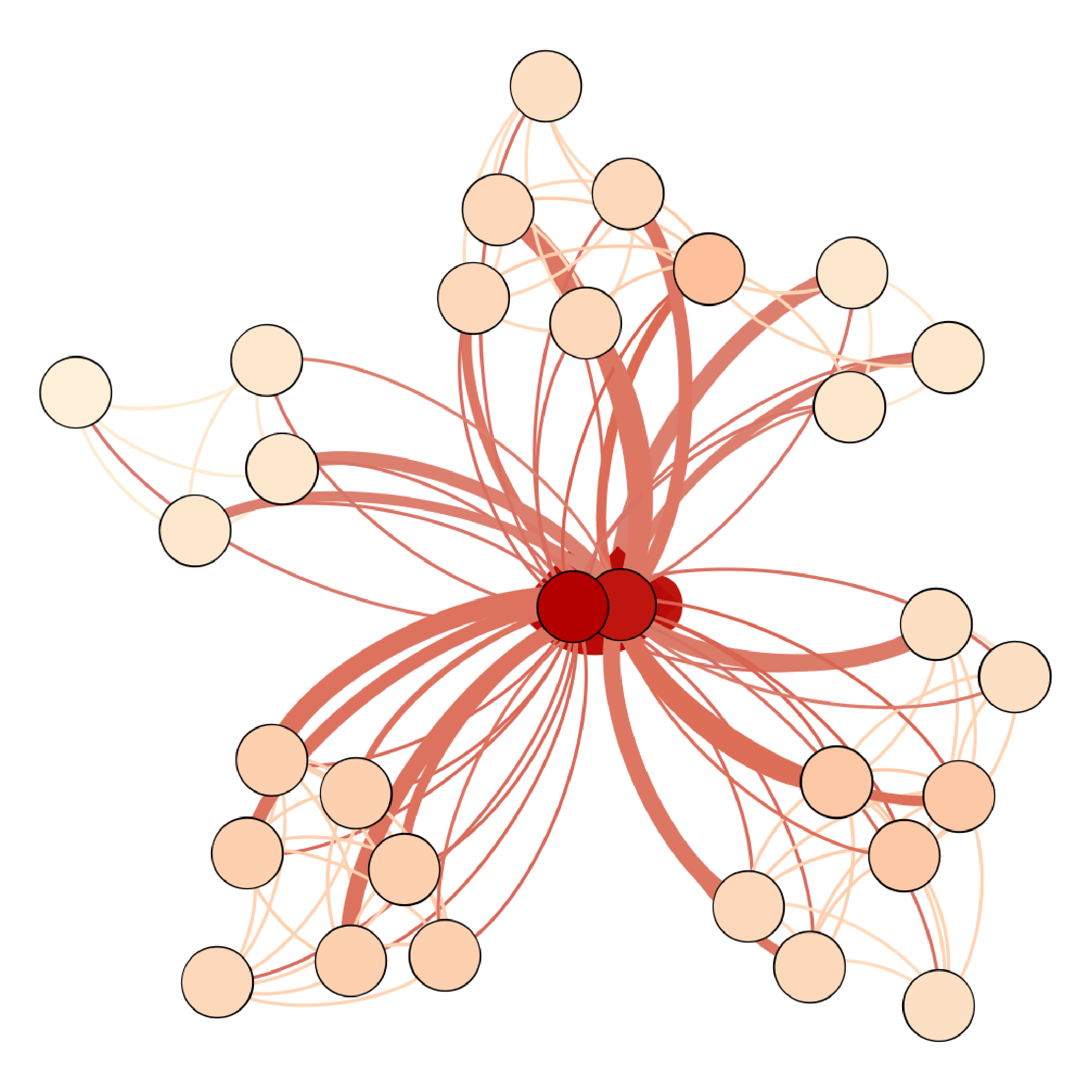}
			
	\end{minipage}	}
	\subfloat[class 6-1]{
		\begin{minipage}[t]{0.25\linewidth}
			\centering
			\includegraphics[width=4cm,height=4cm]{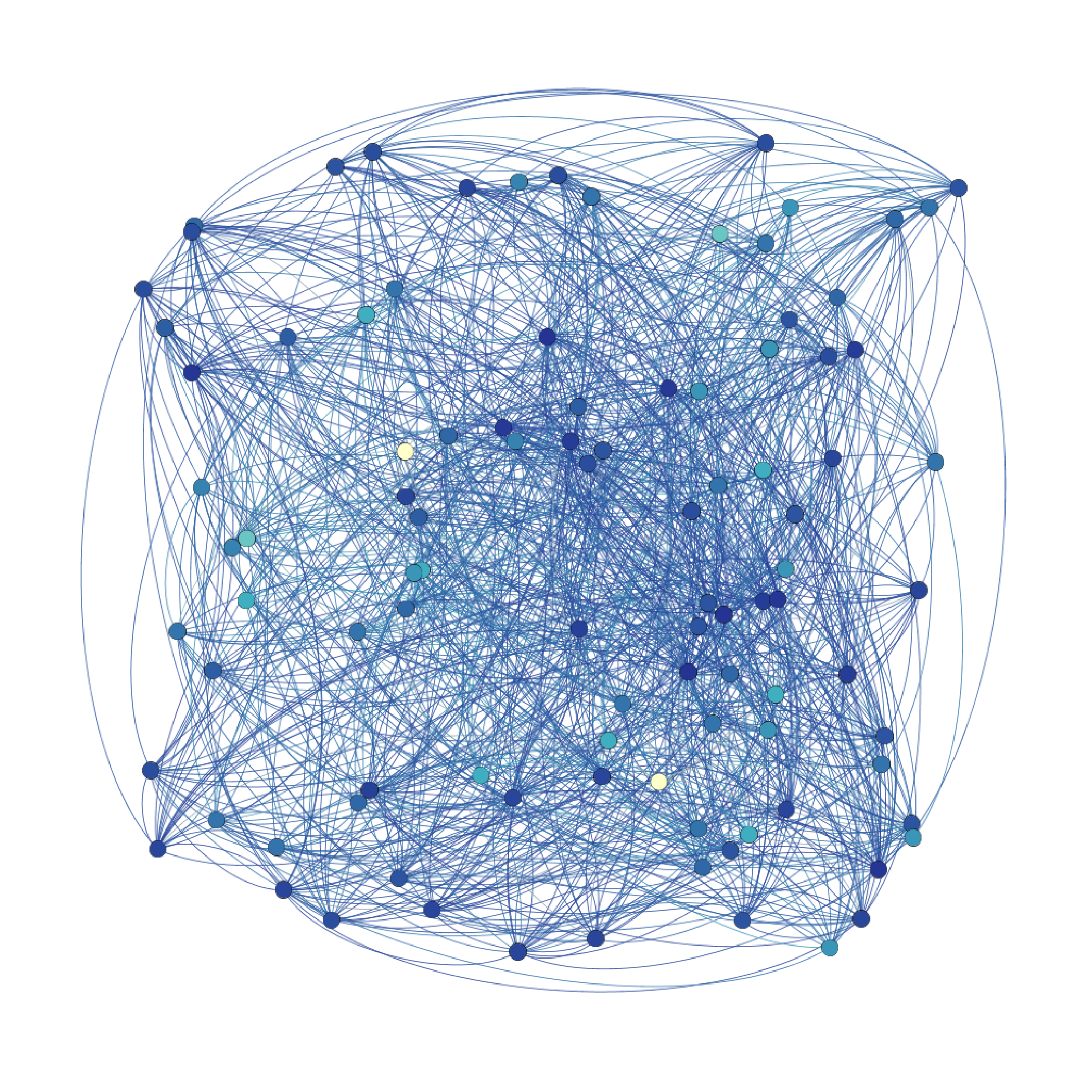}
			
		\end{minipage}
	}
	\subfloat[class 6-2]{
		\begin{minipage}[t]{0.25\linewidth}
			\centering
			\includegraphics[width=4cm,height=4cm]{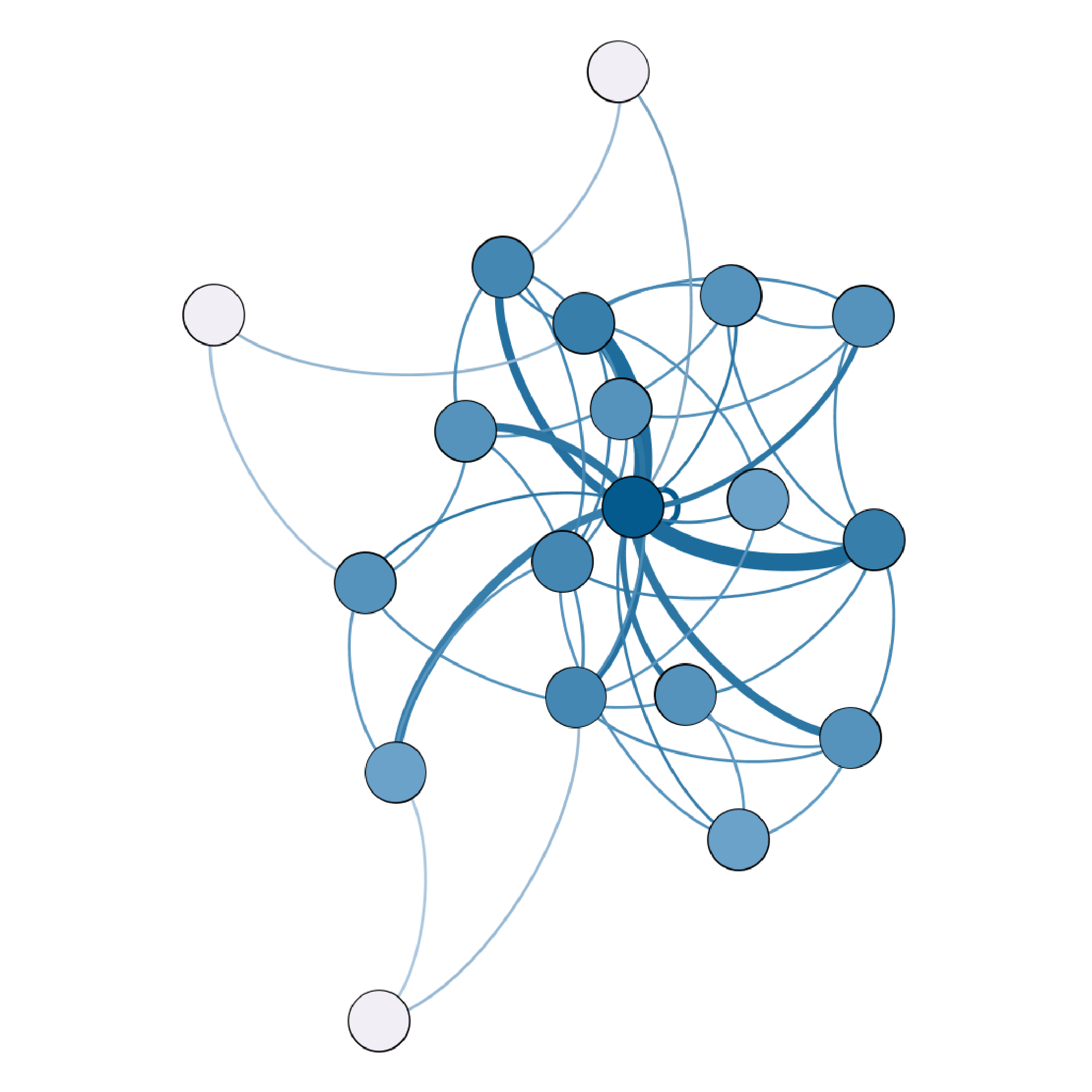}
			
		\end{minipage}
	}
	\centering
	\caption{Example of the MSG data set. class X-Y indicates that the graph is the Y-th example from class X. 
	}
	\label{fig3} 
\end{figure*}

As evident from the Fig. \ref{figdis}, the maximum number of nodes in graphs in MSG is 1,000, and the minimum number of graph's nodes is 4. The difference is approximately $ 10^3 $, which meets the definition of the cross-scale graphs. MSG consists mainly of three peaks: The first peak consists of graphs of nodes between 0 and 200, representing small-scale networks such as small molecular compounds in the real world. The second peak consists of graphs with 500-600 nodes, representing medium-scale complex networks, such as macromolecular networks and brain networks in the real world. The third peak consists of graphs with 900-1000 nodes, representing large graphs, such as social networks in the real world. 

%Fig. \ref {fig3} is some of the visual results of MSG. 

\subsubsection{Baseline}
To answer $Q1$, we select seven advanced methods for comparison:

%WL\cite{shervashidze2011weisfeiler}. The W-L algorithm yields a unique set of features on graphs, meaning that each node on the graph has a unique feature. Therefore, for most irregular graph structures, the obtained features can be used as the criterion of isomorphism of graphs. 
%
%DGK\cite{yanardag2015deep}. DGK applies idea of language modeling to learn the representations of substructures of graphs. 
%
%AWE\cite{ivanov2018anonymous}. AWE uses an anonymized version of random walk to learn the graph representation to represent the graphs and finish the classification tasks.
Set2set\cite{vinyals2015order}. This work presents a read-process-write framework for unordered output data, and proposes an efficient training algorithm (Set2set), which searches for the best possible output sequence during training and prediction. 

GIN\cite{xu2018powerful}. GIN is as powerful as the Weisfeiler-Lehman graph isomorphism test and it achieves state-of-the-art performance.

Diffpool\cite{ying2018hierarchical}. Diffpool uses GNN models to learn a assign matrix which assigns a group of nodes into one node. Its pooling strategy makes the architecture end-to-end trainable. The node drop pooling uses a learnable scoring function to eliminate nodes with low scores. The researchers report that Diffpool has an advantage on big biology data sets in terms of accuracy. 

Nested GCN\cite{zhang2021nested}. Nested graph neural networks (Nested GNN) represents a graph with rooted subgraphs rather than rooted subtrees. Thus, the representations of two graphs that contain many identical subgraphs tend to be similar. It is reported that Nested GNN is highly competitive for graph classification tasks. We use GCN for its basic model. 

DGCNN\cite{zhang2018end}.  DGCNN uses WL algorithm\cite{shervashidze2011weisfeiler} to generate features for nodes and propose a pooling method called SORTPOOL to select the first $m$ nodes to create an equal-size graph which makes it convenient to use the CNN
method to finish the graph classification task. Finally, a CNN is used for the graph classification tasks.

$ G $-Mixup\cite{han2022g}. $ G$-Mixup uses random graph mixing to generate new graphs and, thus, to augment the original data set. Unlike traditional data enhancement methods, $ G$-Mixup can be generated with different topologies. The graph effectively increases the diversity of the data sets. In addition, $ G$-Mixup can also be used in combination with other data enhancement methods to further improve model performance. 

ICL\cite{Zhao2024}. The Information-based Causal Learning (ICL) framework integrates information theory and causality to transform correlation into dependence. This model introduces a mutual information objective to enhance causal features rather than correlational patterns. The paper claims that ICL significantly improves accuracy and robustness in graph classification tasks.

We utilise GCN\cite{kipf2016semi} + Diffpool\cite{ying2018hierarchical} for our ablation study's baseline model. We add wavelet convolution layer (GWC) and spectral-pooling respectively and test which part is more effective. 

Our model and baseline models use the same network structure (for example, layers, activation functions) and the same training hyperparameters (for example, optimizer, learning rate, and gradient clipping). The proportion of training sets, test sets, and validation sets is 8:1:1.
\subsection{Comparison between GSpect and Other Models}

\subsubsection{Classifying Graphs in Open Data Sets}

The performance of GSpect and baseline models on the classification of open data sets are presented in Table \ref{tab1}. GSpect achieves four of the best performance out of five data sets with the average improvements of 1.62\% in classification accuracy. In particular, our model highly improves the performance (by 3.33\%) for the biological macromolecules data set (PROTEINS). The reaon for this is that GWC captures multi-scale messages from a complex structure. Moreover, because every graph should be pooled into the same size, the spectral-pooling method saves most messages during the process of pooling on a larger scale. It must be noted that at the data set IMDB-B, GSpect lags behind G-mixup. This because IMDB-B is a social network data set and has a small number of nodes that have an enormous number of neighbor nodes. GSpect does not consider the effect of these key nodes in graph classification. In addition, G-mixup employs random graph mixing to generate new graphs, ensuring that the mixed graphs retain the fundamental structure of the original graphs, such as connectivity, which may lead to its superior performance in social networks.

\begin{table*}[th]
	\centering
	
	\caption{Comparison experiment in terms of classification accuracy between GSpect and other models on open data sets. The best results are marked in bold font and sub-best results are underlined. }\label{tab1}
	\begin{tabular}{c c c c c  c}
		\hline
		
		\text { Algorithm } & \text { PTC } & \text { MUTAG } & \text { PROTEINS } & \text { D\&D } & \text { IMDB-B } \\
		\hline 
		%		\text { WL } & $ 66.25\pm 0.48 $ & $ 82.31\pm 0.62 $ & $ 74.10\pm 0.62 $ & $ 79.08\pm 0.36 $ & $ 73.40\pm 0.96 $ \\
		%	
		%		\text { DGK } & $ 62.21\pm 2.64 $ & $ 86.96\pm 2.01 $ & $ 74.90\pm 0.85 $ & $ 73.31\pm 1.21 $ & $ 71.51\pm 0.85$ \\
		%		\text { AWE } & $ 67.50\pm 6.71 $ & $ 87.45\pm 4.98 $ & $ 73.73\pm 2.64 $ & $ 72.68\pm 3.52 $ & $ 74.45\pm 2.96$ \\
		%		\hline 
		\text { Set2set} & $ 64.45\pm 5.51 $ & $ 71.90\pm 2.81 $ & $ 74.51\pm 2.26 $ & $ 76.42\pm 3.84 $ & $ 63.91\pm 4.10 $  \\
		\text { GIN } & $ 64.13\pm 8.12 $ & $ 89.40\pm 5.6 $ & $ 76.46\pm 2.88 $ & $ 76.84\pm 3.11 $ & $ 74.66\pm 5.28 $  \\
		
		\text { Diffpool } & $ 66.65\pm 8.57 $ & $ 84.30\pm 2.56 $ & $ \underline{76.96\pm 1.88} $ & $ 78.88\pm 2.87 $ & $ 65.61\pm 1.11 $  \\
		\text { DGCNN } & $ 72.62\pm 1.76 $ & $ 84.66\pm 2.06 $ & $ 70.59\pm 0.34 $ & $ \underline{79.01\pm 0.52} $ & $ 69.90\pm 0.29  $\\
		\text { NestedGCN } & $ 70.26\pm 4.18 $ & $ 73.81\pm 9.70 $ & $ 74.20\pm 2.50 $ & $ 76.53\pm 3.88 $ & $ 73.79\pm 1.18 $ \\
		
		\text { G-mixup } & $ \underline{ 74.41\pm 1.62} $ & $ 87.98\pm 2.49 $ & $ 74.44\pm 1.63 $ & $ 78.61\pm 0.89 $ & $ \pmb{83.84}\pm \pmb{3.20} $ \\
		\text{ICL} & $ 73.02\pm 7.17 $ & $ \underline{89.57\pm 4.06} $ & $ 75.21\pm 2.99 $ & $ 76.15\pm 2.56 $ & $74.59\pm 4.70 $ \\
		\hline 
		\text { \textbf{GSpect} } & $ \pmb{74.90}\pm \pmb{3.53} $ &  $ \pmb{91.11}\pm \pmb{4.68} $ & $ \pmb{80.29}\pm \pmb{2.83} $ & $ \pmb{80.14}\pm \pmb{4.38} $ & $ \underline{74.85\pm  4.00 }$ \\
		\hline
		
	\end{tabular}
\end{table*}

\begin{table*}[ht]
	\centering
	
	\caption{Ablation study between GSpect and other models on open data sets. }\label{tab2}
	\begin{tabular}{c c c c c c c}
		
		\hline \text { Algorithm } & \text { PTC } & \text { MUTAG } & \text { PROTEINS } & \text { D\&D } & \text { IMDB-B }& \text { MSG } \\
		\hline 
		\text { GCN+Diffpool } & $ 66.65\pm 8.57 $ & $ 84.30\pm 2.56 $ & $ 76.96\pm 1.88 $ & $ 77.88\pm 2.87 $ & $ 65.61\pm 1.11 $ & $ 57.14\pm 9.05 $\\
		
		\text { GCN+Spectral-pooling} & $ 67.06\pm 4.96 $ & $ 93.33\pm 5.11 $ & $ 81.18\pm 3.97 $ & $ 81.08\pm 7.02 $ & $ 74.10\pm 2.85  $& $ 71.90\pm 8.73  $\\
		
		\text {GWC+Diffpool } & $ 70.83\pm 8.23 $ & $ 90.55\pm 3.75 $ & $ 78.56\pm 2.64 $ & $ 80.42\pm 3.45 $ & $ 71.86\pm 5.27 $ & $ 70.52\pm 6.48 $
		\\
		\text { \textbf{GSpect} } & $ 74.90\pm 3.53 $ &  $ 91.11\pm 4.68 $ & $ 80.29\pm 2.83 $ & $ 80.14\pm 4.38 $ & $ 74.85\pm  4.00 $ & $ 75.33\pm  7.73 $\\
		\hline
		
	\end{tabular}
\end{table*}

\subsubsection{Classifying Graphs in Cross-scale Data Sets}

\begin{figure}[htb]
	\includegraphics[width=0.5\textwidth]{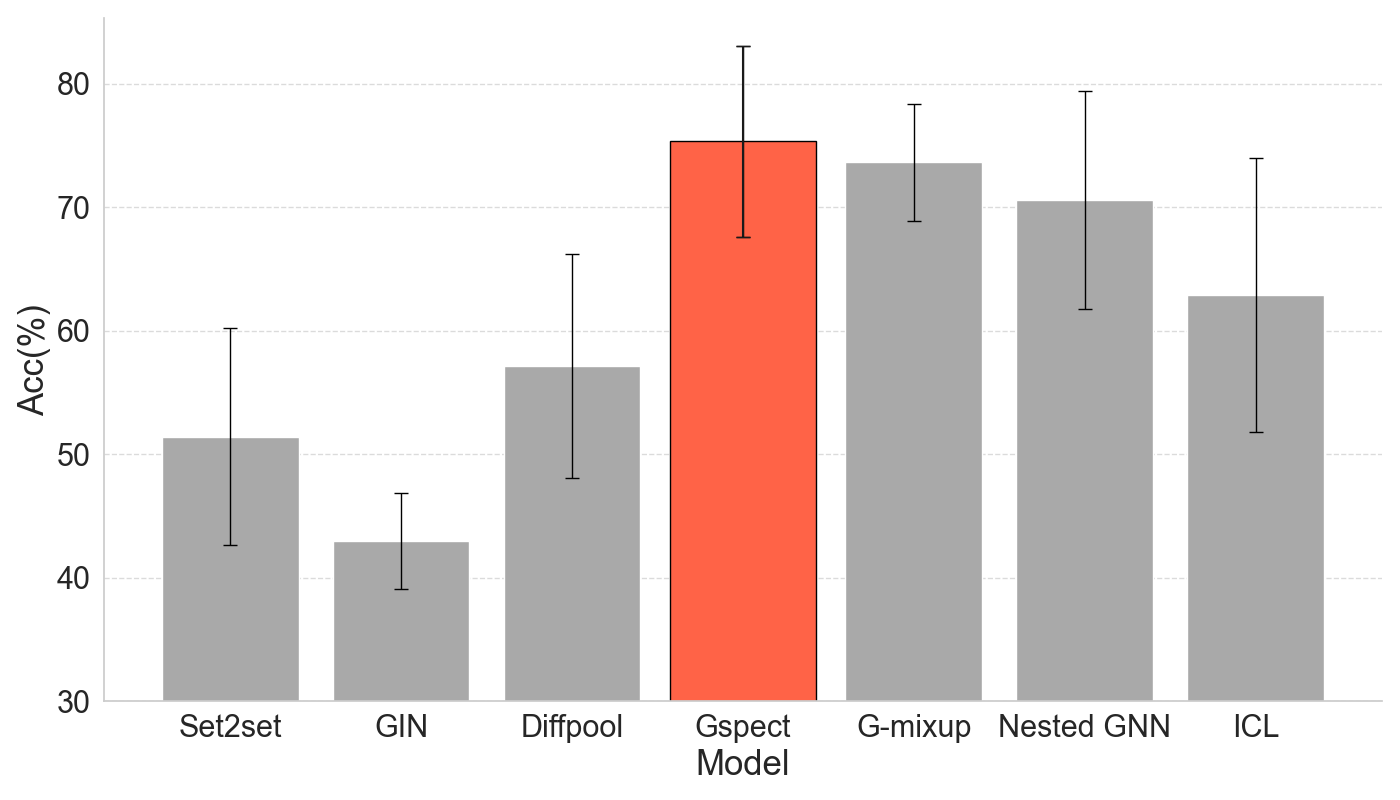}
	\caption{Comparison experiment between GSpect and other models on MSG. } \label{fig5}
\end{figure}

The performance of GSpect and baseline models on the classification of MSG are presented in Fig. \ref{fig5}. We improved the average accuracy by 15.55\% (average difference in accuracy between GSpect and all other methods). There are numerous reasons for this. First, the final GWC layer is composed of multi-scale GWC layers; thus, the advantage of GWC is its ability to capture the information of cross-scale structures in graphs. For the cross-scale graph data set MSG, GWC can better aggregate the structure information and generate graph representations. Second, because the graphs' adjacency matrix is usually different in size, the traditional methods are difficult to pool the graphs. However, these cross-scale graphs in the same class have a similar topology and also have a similar spectrum. Thus, the spectral-pooling method can accomplish the pooling task. 

It must be pointed out that almost all methods yield a large standard deviation, which results in high uncertainty. This is due to a number of reasons. First, collecting cross-scale graph data is difficult and, thus, the sample space of MSG is small (210 samples), thereby leading to large fluctuations. Second, the large variation in graph size (almost $10^3$) leads to the difficulty of classification, which results in a few wrong classification results. 

\begin{figure}[!ht]
	
	\centering  %图片全局居中
	%	\subfigbottomskip=2pt %两行子图之间的行间距
	%	\subfigcapskip=-1pt %设置子图与子标题之间的距离
	\subfloat[Sensitivity analysis of $ F $]{
		\includegraphics[width=\linewidth]{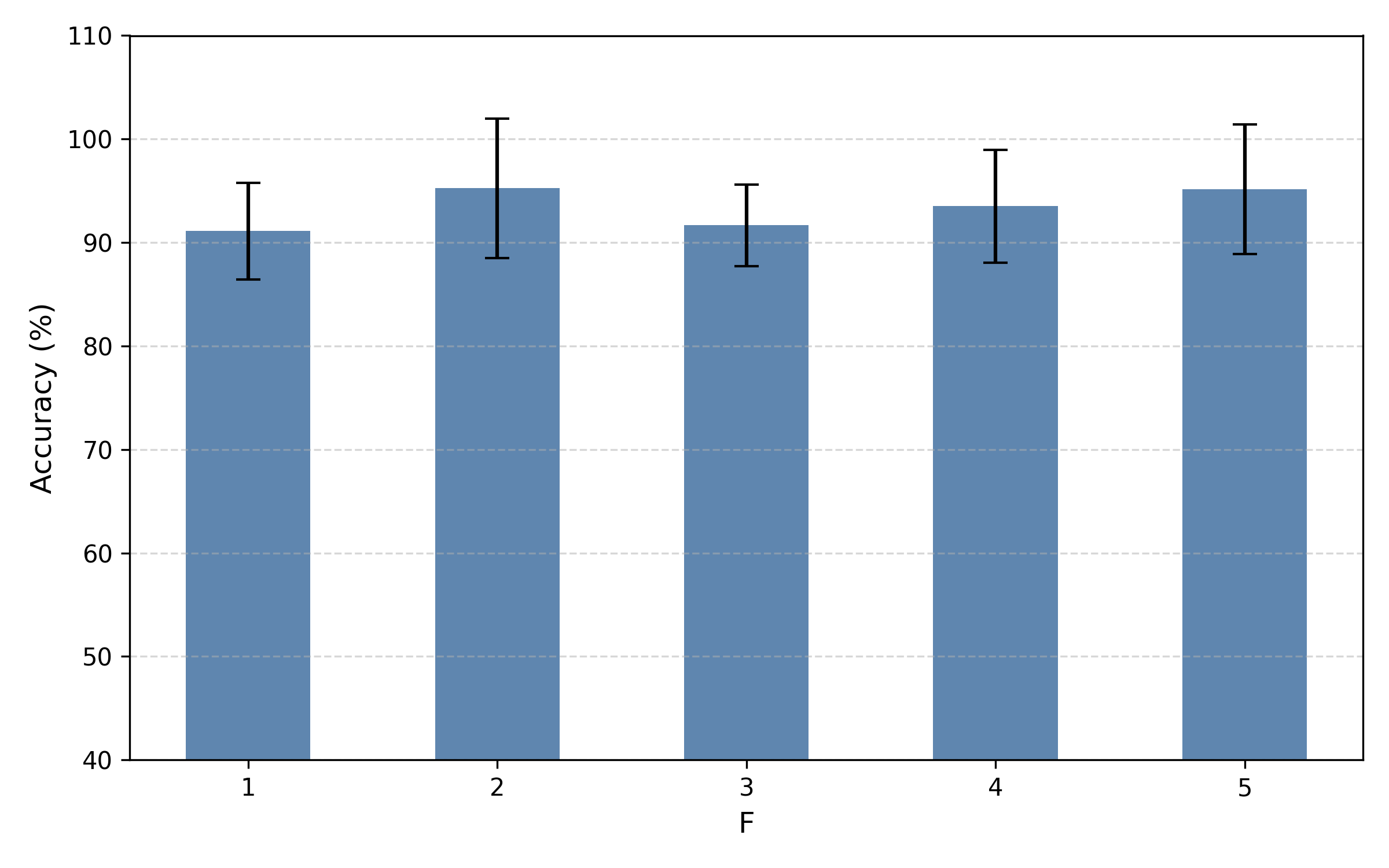}}
	
	\subfloat[Sensitivity analysis of $ M $]{
		\includegraphics[width=\linewidth]{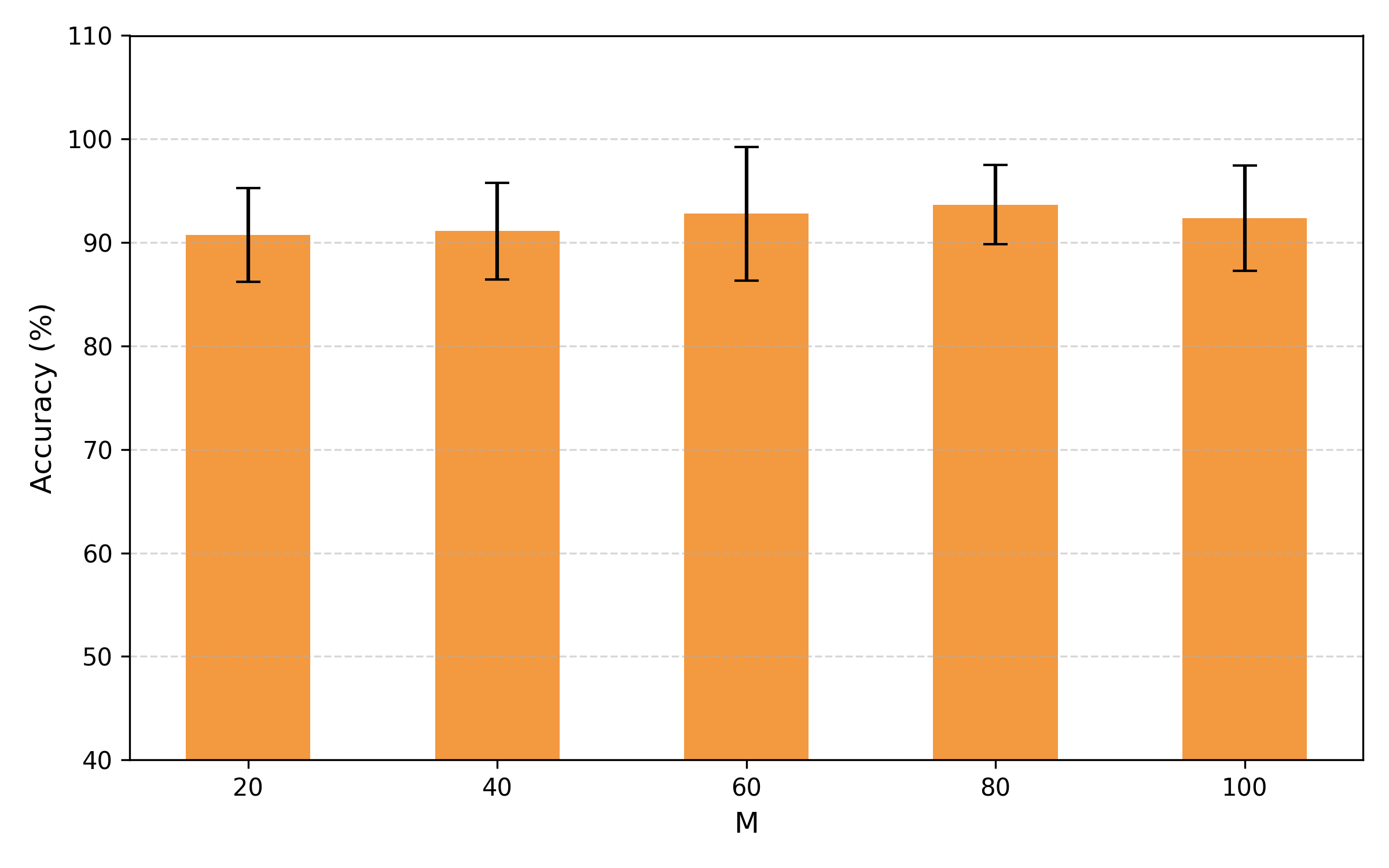}}
	\\	\subfloat[Sensitivity analysis of $ \beta $]{
		\includegraphics[width=\linewidth]{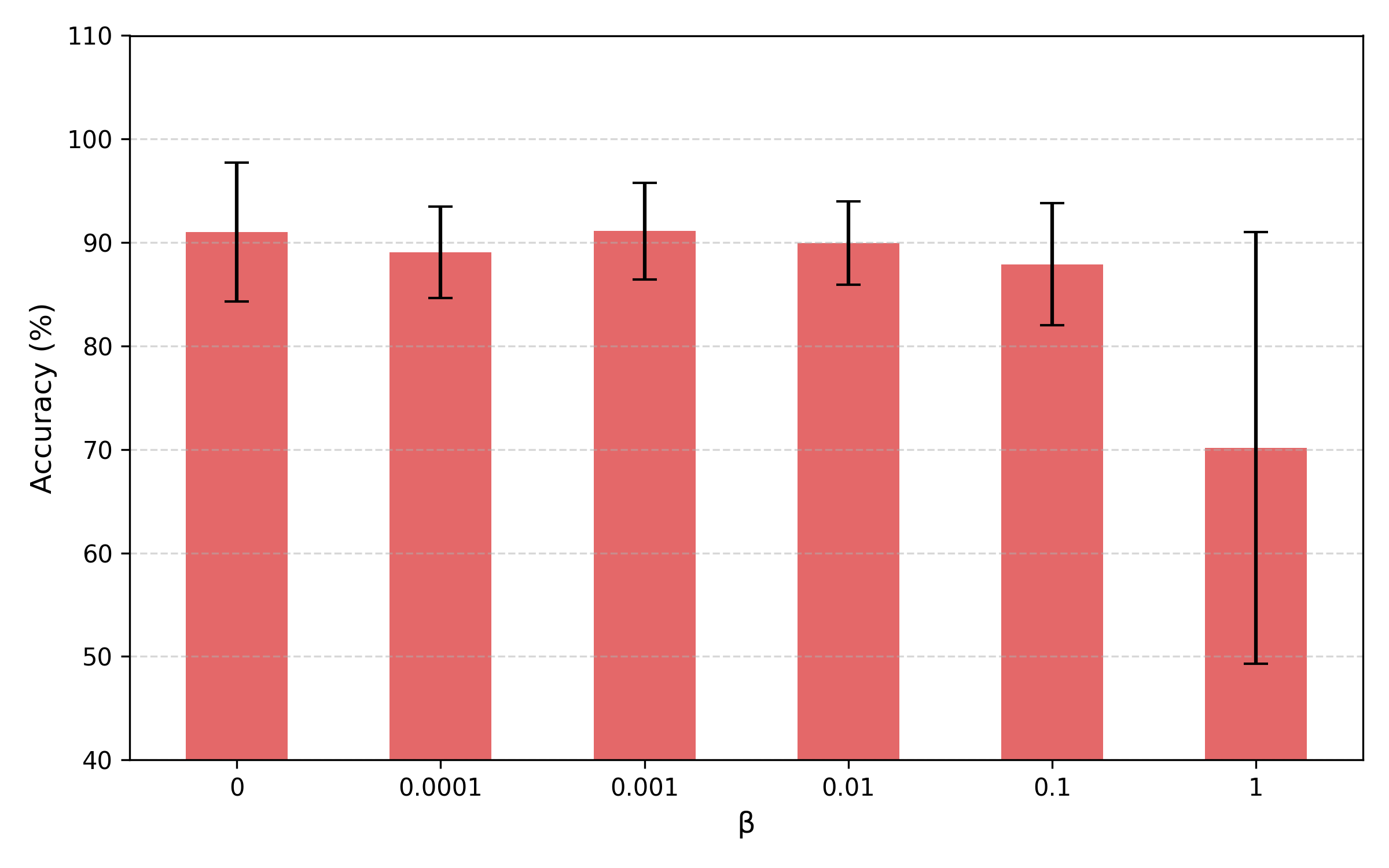}}
	%\quad
	\centering
	\caption{The results of sensitivity analysis.}
	\label{fig6} 
\end{figure}
\subsection{Ablation Study}

To answer $Q2$, we design an ablation study to verify which part of GSpect is significant and why GSpect has better performance. 

Table \ref{tab2} reports the results of ablation study. It is evident that GWC and the spectral-pooling layer improves the performance partly, which proves the effectiveness of GWC and spectral-pooling.

Note that using spectral pooling with the data sets MUTAG and PROTEINS, using spectral-pooling only leads to better performance than GSpect. The reason for this is that the GWC aggregates the multi-scale spectral messages as its output and the spectral-pooling layer filters the redundant messages and generates the principal component representation. However, in this study, we retain the first $ F $-scale wavelet and loss portion of the high-scale messages. For MUTAG and PROTEINS, this method affects the accuracy of classification. Another reason is, unlike most other methods, GWC trains a non-sparse parameter matrix, which may lead to overfitting on these datasets. After removing GWC, the model complexity is reduced, which in turn improves its generalization ability. 

With regard to stability, GWC and spectral-pooling partially increase the standard deviation of classification accuracy, which reduces the stability of the model. This is because these two methods have more learnable parameters which increase the difficulty of optimization and increase the probability of falling into local optimum.   

\subsection{Sensitivity Analysis}

To answer $Q3$, we change the value of hyperparameters and observe the performance of GSpect in the classification accuracy. The experiment is based on MUTAG. Fig. \ref{fig6} presents the results of the sensitivity analysis. According to Fig. \ref{fig6}, we find that GSpect undergoes small changes when the number of Chebyshev polynomials $ M $ and the number of wavelet scale $F$ changes. This result implies that on the basis of maintaining high classification accuracy, researchers can select small $F$ and $M$ to reach lower code execution time. 

However, the classification accuracy reduces sharply when the equilibrium coefficient $\beta$ increases. As Equation \ref{ballence} shows, when $\beta$ is close to 1, $L_p$ plays a leading role in the optimization function. The results reveal that using $L_p$ alone will reduce the performance of GSpect. Thus, researchers need to adjust $\beta$ to ensure that the two optimization function have the same order of magnitude.

\section{Conclusion}
Structure determines function in many systems. As a key method for identifying common structures for functional design and system optimization, cross-scale graph classification is crucial in many aspects such as bioinformatics, drug design, and complex networks.
Considering there is few methods for cross-scale graph classification tasks, we proposed GSpect, an advanced cross-scale graph classification model in this study. 
%Different from existing methods, GSpect use graph wavelet transform for its convolution layer and a spectral filter for its pooling layer. We proposed GWC, a new graph wavelet convolution layer. 
We use the graph wavelet neural network as the convolution layer which improved the performance of obtaining graph-level representations. In addition, we designed the spectral-pooling layer which filters useless messages directly on the spectrum and aggregates the nodes to resize the graph by spectral pooling. Based on the fact that there is few cross-scale graph data sets, we collect data and create the cross-scale data set MSG. We compared this data set with the state-of-the-art ones to prove the superiority of the classification accuracy of GSpect using both open data sets and MSG. Experiments reveal that, on open data sets, GSpect improves the performance of classification accuracy by 1.62\% on average, and for a maximum improvement of 3.33\% on PROTEINS. On MSG, GSpect improves the performance of classification accuracy by 15.55\% on average. Further, we employed an ablation study to observe the improve of accuracy by GWC and spectral-pooling. The results reveal that when we employed them simultaneously, we obtain the best results with regard to graph classification, which proves that it is necessary to use them simultaneously. Further, we conducted the sensitivity analysis to verify the stability of GSpect when there is a change in the hyperparameters. The results reveal that researchers can select a small $F$ and $M$ but need to decide the value of $\beta$ carefully. While GSpect demonstrates excellent performance in cross-scale graph classification tasks, we acknowledge certain limitations inherent in our approach. This work fills the gap of lacking cross-scale graph classification research. Besides, GSpect fills the gap in extant literature regarding a lack of cross-scale graph classification studies and could facilitate application research, for example, predicting the function of protein in accordance with its structure and enabling the selection of appropriate drugs.

%articles~\cite{ref_article1}, an LNCS chapter~\cite{ref_lncs1}, a
%book~\cite{ref_book1}, proceedings without editors~\cite{ref_proc1},
%and a homepage~\cite{ref_url1}. Multiple citations are grouped
%\cite{ref_article1,ref_lncs1,ref_book1},
%\cite{ref_article1,ref_book1,ref_proc1,ref_url1}.

%\section{Acknowledgements} 
%This work was supported by the National Natural Science Foundation of China (72025405, 72088101), the National Social Science Foundation of China (22ZDA102), the Hunan Science and Technology Plan Project (2020TP1013, 2020JJ4673, 2023JJ40685), the Shenzhen Basic Research Project for Development of Science and Technology
%(JCYJ20200109141218676, 202008291726500001), and the Innovation Team Project of Colleges in Guangdong Province (2020KCXTD040). The authors declare that they have no conflict of interest.

%

%\section*{References}
%\nocite{*}
\bibliographystyle{IEEEtran}
\bibliography{ref} %bibfile_name

\begin{thebibliography}{10}
\providecommand{\url}[1]{#1}
\csname url@samestyle\endcsname
\providecommand{\newblock}{\relax}
\providecommand{\bibinfo}[2]{#2}
\providecommand{\BIBentrySTDinterwordspacing}{\spaceskip=0pt\relax}
\providecommand{\BIBentryALTinterwordstretchfactor}{4}
\providecommand{\BIBentryALTinterwordspacing}{\spaceskip=\fontdimen2\font plus
\BIBentryALTinterwordstretchfactor\fontdimen3\font minus
  \fontdimen4\font\relax}
\providecommand{\BIBforeignlanguage}[2]{{%
\expandafter\ifx\csname l@#1\endcsname\relax
\typeout{** WARNING: IEEEtran.bst: No hyphenation pattern has been}%
\typeout{** loaded for the language `#1'. Using the pattern for}%
\typeout{** the default language instead.}%
\else
\language=\csname l@#1\endcsname
\fi
#2}}
\providecommand{\BIBdecl}{\relax}
\BIBdecl

\bibitem{whitford2013proteins}
D.~Whitford, \emph{Proteins: structure and function}.\hskip 1em plus 0.5em
  minus 0.4em\relax {John Wiley \& Sons}, 2013.

\bibitem{lu2016detecting}
X.~Lu, D.~J. Wrathall, P.~R. Sunds{\o}y, M.~Nadiruzzaman, E.~Wetter, A.~Iqbal,
  T.~Qureshi, A.~J. Tatem, G.~S. Canright, K.~Eng{\o}-Monsen \emph{et~al.},
  ``Detecting climate adaptation with mobile network data in bangladesh:
  Anomalies in communication, mobility and consumption patterns during cyclone
  mahasen,'' \emph{{Climatic Change}}, vol. 138, pp. 505--519, 2016.

\bibitem{schadt2009network}
E.~E. Schadt, S.~H. Friend, and D.~A. Shaywitz, ``A network view of disease and
  compound screening,'' \emph{{Nature Reviews Drug Discovery}}, vol.~8, no.~4,
  pp. 286--295, 2009.

\bibitem{wu2014human}
B.~Wu, C.~Yuan, and W.~Hu, ``Human action recognition based on
  context-dependent graph kernels,'' in \emph{{Proceedings of the IEEE
  Conference on Computer Vision and Pattern Recognition}}, 2014, pp.
  2609--2616.

\bibitem{lee2020deep}
J.~B. Lee, X.~Kong, C.~M. Moore, and N.~K. Ahmed, ``Deep parametric model for
  discovering group-cohesive functional brain regions,'' in \emph{{Proceedings
  of the 2020 SIAM International Conference on Data Mining}}.\hskip 1em plus
  0.5em minus 0.4em\relax SIAM, 2020, pp. 631--639.

\bibitem{brown2009chemoinformatics}
N.~Brown, ``Chemoinformatics—an introduction for computer scientists,''
  \emph{{ACM Computing Surveys (CSUR)}}, vol.~41, no.~2, pp. 1--38, 2009.

\bibitem{wen2020convolutional}
J.~Wen, E.~Thibeau-Sutre, M.~Diaz-Melo, J.~Samper-Gonz{\'a}lez, A.~Routier,
  S.~Bottani, D.~Dormont, S.~Durrleman, N.~Burgos, O.~Colliot \emph{et~al.},
  ``Convolutional neural networks for classification of alzheimer's disease:
  Overview and reproducible evaluation,'' \emph{{Medical Image Analysis}},
  vol.~63, p. 101694, 2020.

\bibitem{nikolentzos2021graph}
G.~Nikolentzos, G.~Siglidis, and M.~Vazirgiannis, ``Graph kernels: A survey,''
  \emph{{Journal of Artificial Intelligence Research}}, vol.~72, pp. 943--1027,
  2021.

\bibitem{kipf2016semi}
T.~N. Kipf and M.~Welling, ``Semi-supervised classification with graph
  convolutional networks,'' \emph{arXiv preprint arXiv:1609.02907}, 2016.

\bibitem{zhuang2018dual}
C.~Zhuang and Q.~Ma, ``Dual graph convolutional networks for graph-based
  semi-supervised classification,'' in \emph{{Proceedings of the 2018 World
  Wide Web Conference}}, 2018, pp. 499--508.

\bibitem{wu2020comprehensive}
Z.~Wu, S.~Pan, F.~Chen, G.~Long, C.~Zhang, and S.~Y. Philip, ``A comprehensive
  survey on graph neural networks,'' \emph{{IEEE Transactions on Neural
  Networks and Learning Systems}}, vol.~32, no.~1, pp. 4--24, 2020.

\bibitem{dobson2003distinguishing}
P.~D. Dobson and A.~J. Doig, ``Distinguishing enzyme structures from
  non-enzymes without alignments,'' \emph{{Journal of Molecular Biology}}, vol.
  330, no.~4, pp. 771--783, 2003.

\bibitem{yanardag2015deep}
P.~Yanardag and S.~Vishwanathan, ``Deep graph kernels,'' in \emph{{Proceedings
  of the 21th ACM SIGKDD International Conference on Knowledge Discovery and
  Data Mining}}, 2015, pp. 1365--1374.

\bibitem{debnath1991structure}
A.~K. Debnath, R.~L. Lopez~de Compadre, G.~Debnath, A.~J. Shusterman, and
  C.~Hansch, ``Structure-activity relationship of mutagenic aromatic and
  heteroaromatic nitro compounds. correlation with molecular orbital energies
  and hydrophobicity,'' \emph{{Journal of Medicinal Chemistry}}, vol.~34,
  no.~2, pp. 786--797, 1991.

\bibitem{silva2019novo}
D.-A. Silva, S.~Yu, U.~Y. Ulge, J.~B. Spangler, K.~M. Jude,
  C.~Lab{\~a}o-Almeida, L.~R. Ali, A.~Quijano-Rubio, M.~Ruterbusch, I.~Leung
  \emph{et~al.}, ``De novo design of potent and selective mimics of il-2 and
  il-15,'' \emph{{Nature}}, vol. 565, no. 7738, pp. 186--191, 2019.

\bibitem{cao2020novo}
L.~Cao, I.~Goreshnik, B.~Coventry, J.~B. Case, L.~Miller, L.~Kozodoy, R.~E.
  Chen, L.~Carter, A.~C. Walls, Y.-J. Park \emph{et~al.}, ``De novo design of
  picomolar sars-cov-2 miniprotein inhibitors,'' \emph{{Science}}, vol. 370,
  no. 6515, pp. 426--431, 2020.

\bibitem{glasgow2019computational}
A.~A. Glasgow, Y.-M. Huang, D.~J. Mandell, M.~Thompson, R.~Ritterson, A.~L.
  Loshbaugh, J.~Pellegrino, C.~Krivacic, R.~A. Pache, K.~A. Barlow
  \emph{et~al.}, ``Computational design of a modular protein sense-response
  system,'' \emph{{Science}}, vol. 366, no. 6468, pp. 1024--1028, 2019.

\bibitem{hsia2016design}
Y.~Hsia, J.~B. Bale, S.~Gonen, D.~Shi, W.~Sheffler, K.~K. Fong, U.~Nattermann,
  C.~Xu, P.-S. Huang, R.~Ravichandran \emph{et~al.}, ``Design of a hyperstable
  60-subunit protein icosahedron,'' \emph{{Nature}}, vol. 535, no. 7610, pp.
  136--139, 2016.

\bibitem{redfern2008exploring}
O.~C. Redfern, B.~Dessailly, and C.~A. Orengo, ``Exploring the structure and
  function paradigm,'' \emph{{Current Opinion in Structural Biology}}, vol.~18,
  no.~3, pp. 394--402, 2008.

\bibitem{du2010clustering}
K.-L. Du, ``Clustering: A neural network approach,'' \emph{{Neural Networks}},
  vol.~23, no.~1, pp. 89--107, 2010.

\bibitem{tan2003general}
S.-D. Tan, ``A general s-domain hierarchical network reduction algorithm,'' in
  \emph{{ICCAD-2003. International Conference on Computer Aided Design (IEEE
  Cat. No. 03CH37486)}}.\hskip 1em plus 0.5em minus 0.4em\relax IEEE, 2003, pp.
  650--657.

\bibitem{slota2016complex}
G.~M. Slota, K.~Madduri, and S.~Rajamanickam, ``Complex network partitioning
  using label propagation,'' \emph{{SIAM Journal on Scientific Computing}},
  vol.~38, no.~5, pp. S620--S645, 2016.

\bibitem{liu2022size}
Z.~Liu, Q.~Mao, C.~Liu, Y.~Fang, and J.~Sun, ``On size-oriented long-tailed
  graph classification of graph neural networks,'' in \emph{Proceedings of the
  ACM Web Conference 2022}, 2022, pp. 1506--1516.

\bibitem{hammond2011wavelets}
D.~K. Hammond, P.~Vandergheynst, and R.~Gribonval, ``Wavelets on graphs via
  spectral graph theory,'' \emph{{Applied and Computational Harmonic
  Analysis}}, vol.~30, no.~2, pp. 129--150, 2011.

\bibitem{shuman2013emerging}
D.~I. Shuman, S.~K. Narang, P.~Frossard, A.~Ortega, and P.~Vandergheynst, ``The
  emerging field of signal processing on graphs: Extending high-dimensional
  data analysis to networks and other irregular domains,'' \emph{{IEEE Signal
  Processing Magazine}}, vol.~30, no.~3, pp. 83--98, 2013.

\bibitem{defferrard2016convolutional}
M.~Defferrard, X.~Bresson, and P.~Vandergheynst, ``Convolutional neural
  networks on graphs with fast localized spectral filtering,'' \emph{{Advances
  in Neural Information Processing Systems}}, vol.~29, 2016.

\bibitem{tremblay2014graph}
N.~Tremblay and P.~Borgnat, ``Graph wavelets for multiscale community mining,''
  \emph{IEEE Transactions on Signal Processing}, vol.~62, no.~20, pp.
  5227--5239, 2014.

\bibitem{xu2019graph}
B.~Xu, H.~Shen, Q.~Cao, Y.~Qiu, and X.~Cheng, ``Graph wavelet neural network,''
  \emph{arXiv preprint arXiv:1904.07785}, 2019.

\bibitem{coifman2006diffusion}
R.~R. Coifman and M.~Maggioni, ``Diffusion wavelets,'' \emph{Applied and
  computational harmonic analysis}, vol.~21, no.~1, pp. 53--94, 2006.

\bibitem{donnat2018learning}
C.~Donnat, M.~Zitnik, D.~Hallac, and J.~Leskovec, ``Learning structural node
  embeddings via diffusion wavelets,'' in \emph{Proceedings of the 24th ACM
  SIGKDD international conference on knowledge discovery \& data mining}, 2018,
  pp. 1320--1329.

\bibitem{bruna2013spectral}
J.~Bruna, W.~Zaremba, A.~Szlam, and Y.~LeCun, ``Spectral networks and locally
  connected networks on graphs,'' \emph{arXiv preprint arXiv:1312.6203}, 2013.

\bibitem{erdHos1960evolution}
P.~Erd{\H{o}}s, A.~R{\'e}nyi \emph{et~al.}, ``On the evolution of random
  graphs,'' \emph{{Publication of the Mathematical Institute of the Hungarian
  Academy of Sciences}}, vol.~5, no.~1, pp. 17--60, 1960.

\bibitem{watts1998collective}
D.~J. Watts and S.~H. Strogatz, ``Collective dynamics of
  ‘small-world’networks,'' \emph{{Nature}}, vol. 393, no. 6684, pp.
  440--442, 1998.

\bibitem{barabasi1999emergence}
A.-L. Barab{\'a}si and R.~Albert, ``Emergence of scaling in random networks,''
  \emph{{Science}}, vol. 286, no. 5439, pp. 509--512, 1999.

\bibitem{ma2021deep}
G.~Ma, N.~K. Ahmed, T.~L. Willke, and P.~S. Yu, ``Deep graph similarity
  learning: A survey,'' \emph{{Data Mining and Knowledge Discovery}}, vol.~35,
  pp. 688--725, 2021.

\bibitem{shervashidze2011weisfeiler}
N.~Shervashidze, P.~Schweitzer, E.~J. Van~Leeuwen, K.~Mehlhorn, and K.~M.
  Borgwardt, ``Weisfeiler-lehman graph kernels.'' \emph{{Journal of Machine
  Learning Research}}, vol.~12, no.~9, 2011.

\bibitem{al2019ddgk}
R.~Al-Rfou, B.~Perozzi, and D.~Zelle, ``Ddgk: Learning graph representations
  for deep divergence graph kernels,'' in \emph{{The World Wide Web
  Conference}}, 2019, pp. 37--48.

\bibitem{gilmer2017neural}
J.~Gilmer, S.~S. Schoenholz, P.~F. Riley, O.~Vinyals, and G.~E. Dahl, ``Neural
  message passing for quantum chemistry,'' in \emph{{International Conference
  on Machine Learning}}.\hskip 1em plus 0.5em minus 0.4em\relax PMLR, 2017, pp.
  1263--1272.

\bibitem{velivckovic2017graph}
P.~Veli{\v{c}}kovi{\'c}, G.~Cucurull, A.~Casanova, A.~Romero, P.~Lio, and
  Y.~Bengio, ``Graph attention networks,'' \emph{arXiv preprint
  arXiv:1710.10903}, 2017.

\bibitem{xu2018powerful}
K.~Xu, W.~Hu, J.~Leskovec, and S.~Jegelka, ``{How Powerful are Graph Neural
  Networks?}'' \emph{arXiv preprint arXiv:1810.00826}, 2018.

\bibitem{duvenaud2015convolutional}
D.~K. Duvenaud, D.~Maclaurin, J.~Iparraguirre, R.~Bombarell, T.~Hirzel,
  A.~Aspuru-Guzik, and R.~P. Adams, ``Convolutional networks on graphs for
  learning molecular fingerprints,'' \emph{{Advances in Neural Information
  Processing Systems}}, vol.~28, 2015.

\bibitem{ma2019graph}
Y.~Ma, S.~Wang, C.~C. Aggarwal, and J.~Tang, ``Graph convolutional networks
  with eigenpooling,'' in \emph{{Proceedings of the 25th ACM SIGKDD
  International Conference on Knowledge Discovery \& Data Mining}}, 2019, pp.
  723--731.

\bibitem{bianchi2020spectral}
F.~M. Bianchi, D.~Grattarola, and C.~Alippi, ``Spectral clustering with graph
  neural networks for graph pooling,'' in \emph{{International Conference on
  Machine Learning}}.\hskip 1em plus 0.5em minus 0.4em\relax PMLR, 2020, pp.
  874--883.

\bibitem{ying2018hierarchical}
Z.~Ying, J.~You, C.~Morris, X.~Ren, W.~Hamilton, and J.~Leskovec,
  ``Hierarchical graph representation learning with differentiable pooling,''
  \emph{{Advances in Neural Information Processing Systems}}, vol.~31, 2018.

\bibitem{yahia2022wavelet}
S.~Yahia, S.~Said, and M.~Zaied, ``Wavelet extreme learning machine and deep
  learning for data classification,'' \emph{{Neurocomputing}}, vol. 470, pp.
  280--289, 2022.

\bibitem{behmanesh2022geometric}
M.~Behmanesh, P.~Adibi, S.~M.~S. Ehsani, and J.~Chanussot, ``Geometric
  multimodal deep learning with multiscaled graph wavelet convolutional
  network,'' \emph{{IEEE Transactions on Neural Networks and Learning
  Systems}}, 2022.

\bibitem{arfken2011mathematical}
G.~B. Arfken, H.~J. Weber, and F.~E. Harris, \emph{Mathematical methods for
  physicists: a comprehensive guide}.\hskip 1em plus 0.5em minus 0.4em\relax
  {Academic Press}, 2011.

\bibitem{liu2022graph}
C.~Liu, Y.~Zhan, C.~Li, B.~Du, J.~Wu, W.~Hu, T.~Liu, and D.~Tao, ``Graph
  pooling for graph neural networks: Progress, challenges, and opportunities,''
  \emph{arXiv preprint arXiv:2204.07321}, 2022.

\bibitem{toivonen2003statistical}
H.~Toivonen, A.~Srinivasan, R.~D. King, S.~Kramer, and C.~Helma, ``Statistical
  evaluation of the predictive toxicology challenge 2000--2001,''
  \emph{{Bioinformatics}}, vol.~19, no.~10, pp. 1183--1193, 2003.

\bibitem{borgwardt2005protein}
K.~M. Borgwardt, C.~S. Ong, S.~Sch{\"o}nauer, S.~Vishwanathan, A.~J. Smola, and
  H.-P. Kriegel, ``Protein function prediction via graph kernels,''
  \emph{{Bioinformatics}}, vol.~21, no. suppl\_1, pp. i47--i56, 2005.

\bibitem{vinyals2015order}
O.~Vinyals, S.~Bengio, and M.~Kudlur, ``Order matters: Sequence to sequence for
  sets,'' \emph{arXiv preprint arXiv:1511.06391}, 2015.

\bibitem{zhang2021nested}
M.~Zhang and P.~Li, ``Nested graph neural networks,'' \emph{arXiv preprint
  arXiv:2110.13197}, 2021.

\bibitem{zhang2018end}
M.~Zhang, Z.~Cui, M.~Neumann, and Y.~Chen, ``An end-to-end deep learning
  architecture for graph classification,'' in \emph{{Proceedings of the AAAI
  Conference on Artificial Intelligence}}, vol.~32, no.~1, 2018.

\bibitem{han2022g}
X.~Han, Z.~Jiang, N.~Liu, and X.~Hu, ``G-mixup: Graph data augmentation for
  graph classification,'' in \emph{{International Conference on Machine
  Learning}}.\hskip 1em plus 0.5em minus 0.4em\relax PMLR, 2022, pp.
  8230--8248.

\bibitem{Zhao2024}
\BIBentryALTinterwordspacing
Z.~Zhao, P.~Wang, H.~Wen, Y.~Zhang, Z.~Zhou, and Y.~Wang, ``A twist for graph
  classification: Optimizing causal information flow in graph neural
  networks,'' \emph{Proceedings of the AAAI Conference on Artificial
  Intelligence}, vol.~38, no.~15, pp. 17\,042--17\,050, Mar. 2024. [Online].
  Available: \url{https://ojs.aaai.org/index.php/AAAI/article/view/29648}
\BIBentrySTDinterwordspacing

\end{thebibliography}
\clearpage
\appendices  

\section{Stability of GSpect}  \label{sta}
In this section, we establish the stability of GSpect. Given that the model comprises two serially connected modules (GWC layer and spectral pooling layer), we independently demonstrate the stability of each module. The proof strategy is as follows: first, we prove that the model satisfies the Lipschitz continuity condition, and then we prove the model's stability.
\subsection{Lipschitz Continuity of GSpect}
Lipschitz continuity is a prerequisite for stability. The definition of Lipschitz continuity is given below. 

\begin{definition}
A function \( f: \mathbb{R}^n \rightarrow \mathbb{R}^m \) is said to be Lipschitz continuous if there exists a constant \( K \geq 0 \), known as the Lipschitz constant, such that for all \( x, y \in \mathbb{R}^n \), the following inequality holds:  
\begin{equation}
	\| f(x) - f(y) \| \leq K \| x - y \|. 
\end{equation}
\end{definition}

Lipschitz continuity implies that the function \( f \) does not oscillate too wildly; small changes in the input \( x \) result in small changes in the output \( f(x) \). Therefore, Lipschitz continuity is crucial for ensuring predictable behavior of dynamical systems and for the convergence of numerical methods. 

In this section, we first establish the Lipschitz continuity of GWC and then establish the Lipschitz continuity of spectral-pooling.
\subsubsection{Lipschitz Continuity of GWC}
We first prove that the graph wavelet transform $\Psi_f$ is Lipschitz continuous. This proof is based on the definition of the graph wavelet transform given in equation \ref{psi}. Then we prove the Lipschitz continuity of GWC. The proof of $\Psi_f$'s Lipschitz continuity is given below. 

\begin{proof} 
We first discuss the continuity of $\Psi_f$, which is essential for GWC's proof part and is helpful for understanding the properties of the graph wavelet transform. Recall that the graph wavelet transform is defined as:
	\begin{equation} \Psi_f = \frac{1}{2} \left(c_{0,f} + \sum_{i=1}^M c_{i,f} T_i(\tilde{L})\right) ,
	\end{equation}
where $c_{i,f} = 2e^{-f} J_i(-f)$, $\tilde{L}$ is the normalized Laplacian matrix, $T_i$ are Chebyshev polynomials, and $J_i$ are Bessel functions of the first kind.
	
Next, we will discuss the continuity of $\Psi_f$'s individual components. First, the constant term $c_{0,f}$ is trivially Lipschitz continuous. Besides, notice that:
\begin{equation}  
	\begin{aligned}  
		|c_{i,f}| &= |2e^{-f} J_i(-f)| \\
		&\leq 2e^{-f} \cdot \max|J_i(-f)|,
	\end{aligned}  
\end{equation}  
which illustrates that $|c_{i,f}|$ is bounded.
Additionally, as $f$ is a fixed scale parameter and Bessel functions are bounded, $c_{i,f}$ is bounded. Finally, the Chebyshev polynomials $T_i(x)$ are Lipschitz continuous on $[-1,1]$, with Lipschitz constants related to their degree $i$. Notice the facts above, $\Psi_f$ is a finite linear combination of Lipschitz continuous functions, which preserves its Lipschitz continuity.

Then we prove the $\Psi_f$'s Lipschitz continuity itself. Let $K_{\Psi,i}$ be the Lipschitz constant of $c_{i,f} T_i(\tilde{L})$. Then the Lipschitz constant $K_{\Psi}$ of $\Psi_f$ can be expressed as:
	
	\begin{equation} 
		K_{\Psi} = \frac{1}{2} \left(|c_{0,f}| + \sum_{i=1}^M K_{\Psi,i}\right) .
	\end{equation}
	Therefore, for any two graphs $G_1$ and $G_2$ with corresponding feature vectors $x_1$ and $x_2$:
	\begin{equation} 
		\|\Psi_f(x_1) - \Psi_f(x_2)\| \leq K_{\Psi} \|x_1 - x_2\|, 
	\end{equation}
	Up to this point, we have proven the Lipschitz continuity of $\Psi_f$.
	
	\end{proof}
	
	Then, we will prove the Lipschitz continuity of the GWC layer.
	\begin{proof}
	Let $F_{GWC}$ denote the operation of the GWC layer. Recall the GWC operation:
		\begin{equation}
		H^{(k+1)}_{n\times l,f} = \sigma(\Psi_{n\times n,f} \Theta_{n\times n} \Psi^{-1}_{n\times n,f} H^k_{n\times l,f} + \hbar_{n\times l}).
		\end{equation}
	For any two inputs $H^k_1$ and $H^k_2$, we need to prove that there exists a constant $K_1$ such that:
		\begin{equation}
			\|F_{GWC}(H^k_1) - F_{GWC}(H^k_2)\| \leq K_1 \|H^k_1 - H^k_2\|.
		\end{equation}
	First, note that $\sigma$ is typically chosen to be a Lipschitz continuous function (such as ReLU or sigmoid) with Lipschitz constant $L_\sigma$.
		
	Let $A = \Psi_{n\times n,f} \Theta_{n\times n} \Psi^{-1}_{n\times n,f}$. Then the bound of $\|F_{GWC}(H^k_1) - F_{GWC}(H^k_2)\|$ is:
	\begin{equation}  
	\begin{aligned}
			\|F_{GWC}(H^k_1) - F_{GWC}(H^k_2)\| &= \|\sigma(AH^k_1 + \hbar) - \sigma(AH^k_2 + \hbar)\| \\
			&\leq L_\sigma \|AH^k_1 - AH^k_2\| \\
			&= L_\sigma \|A(H^k_1 - H^k_2)\| \\
			&\leq L_\sigma \|A\| \|H^k_1 - H^k_2\|.
	\end{aligned}
	\end{equation}	
	Let $K_1 = L_\sigma \|A\|$. Then:
    \begin{equation}  
		\|F_{GWC}(H^k_1) - F_{GWC}(H^k_2)\| \leq K_1 \|H^k_1 - H^k_2\|.  
	\end{equation} 
	In terms of the components in $K_1$, it has been previously proven that \(\Psi_f\) is Lipschitz continuous, and after training, \(\Theta_{n \times n}\) becomes a constant. Therefore, the GWC layer is Lipschitz continuous with Lipschitz constant $K_1 = L_\sigma \|\Psi_{n\times n,f} \Theta_{n\times n} \Psi^{-1}_{n\times n,f}\|$.
	\end{proof}
	
	Then we continue to prove the Lipschitz continuity of spectral-pooling layer.
\subsubsection{Lipschitz Continuity of Spectral-pooling Layer}
We note the spectral pooling layer as $F_{sp}$. For any inputs $X_1$ and $X_2$, we have:
		\begin{equation}
\begin{aligned}
	\|F_{sp}(X_1) - F_{sp}(X_2)\| &= \|S^T X_1 S - S^T X_2 S\| \\
	&= \|S^T (X_1 - X_2) S\|.
\end{aligned}
	\end{equation}
Noticing the properties of matrix norms:
\begin{equation}
	\|S^T (X_1 - X_2) S\| \leq \|S^T\| \|X_1 - X_2\| \|S\|,
\end{equation}
where $\|\cdot\|$ denotes the spectral norm (maximum singular value) of the matrix.

Let $K_2 = \|S^T\| \|S\|$, then:
\begin{equation}
	\|F(X_1) - F(X_2)\| \leq K_2 \|X_1 - X_2\|.
\end{equation}
Since $S$ is a fixed learned matrix, $K_2$ is a constant.

Hence, we have proved the Lipschitz continuity of spectral-pooling layer. 

\subsection{Stability}

Stability is a fundamental concept in the analysis of dynamical systems. First we introduce the definition of stability. 
\begin{definition}
A solution \( x(t) \) of a differential equation is said to be stable, if:
   
	$\forall \epsilon > 0, \exists \delta > 0 \text{ such that if } \|x(t_0) - x_0\| < \delta, \text{ then } \|x(t) - x_0\| < \epsilon \text{ for all } t \geq t_0$.   
\end{definition}

This implies that small changes in the initial condition \( x(t_0) \) result in small changes in the solution \( x(t) \).  

%\end{document}  
Next, we will prove the stability of GSpect. The proof is divided into two parts: first, we demonstrate the stability of GWC layer, and then we prove the stability of spectral pooling layer.
\subsubsection{Stability of GWC Layer}
%Consider an input graph $G$ subject to a small perturbation $\delta$. Since the Lipschitz continuity of GSpect has been proven earlier, we will denote its continuity coefficient as constant $K_{total}$ here. Then we have:
%\begin{equation}
%	|GSpect(G + \delta) - GSpect(G)| \leq K_{total} |\delta|
%\end{equation}
%
%This inequality demonstrates that the output change is bounded by the magnitude of the input perturbation, scaled by the model's Lipschitz constant.

Based on the definition of the GWC layer in the original text:
\begin{equation}
	H^{(k+1)}_{n\times l,f} = \sigma(\Psi_{n\times n,f} \Theta_{n\times n} \Psi^{-1}_{n\times n,f} H^k_{n\times l,f} + \hbar_{n\times l}),
\end{equation}
we simply denote it as:
\begin{equation}
	F_{GWC}(X) = \Psi \Theta \Psi^{-1} X + \hbar.
\end{equation}
%For the input perturbation $X + \delta$, we have:
%\begin{equation}  
%	\begin{aligned}  
%		GWC(X + \delta) &= \Psi \Theta \Psi^{-1} (X + \delta) \\
%		&= \Psi \Theta \Psi^{-1} X + \Psi \Theta \Psi^{-1} \delta  
%	\end{aligned}  
%\end{equation}
%Thus, the difference in GWC outputs can be written as:
%
%\begin{equation}
%	GWC(X + \delta) - GWC(X) = \Psi \Theta \Psi^{-1} \delta
%\end{equation}
%
%For the first term, we can obtain:
%
%\begin{equation}
%	|GWC(X + \delta) - GWC(X)| = |\Psi \Theta \Psi^{-1} \delta|
%\end{equation}
%
%Using the properties of matrix multiplication, we can apply the triangle inequality:
%
%\begin{equation}
%	|\Psi \Theta \Psi^{-1} \delta| \leq |\Psi| |\Theta| |\Psi^{-1}| |\delta|
%\end{equation}

Notably, when there is a change $\delta$ on the input, we express the change in GWC as:
\begin{equation}  
	\begin{aligned}   
		&F_{GWC}(X + \delta) - F_{GWC}(X) \\ 
		&= \Psi \Theta \Psi^{-1} \delta + \sum_{s} \left( U g_s(\Lambda + \Delta\Lambda) U^T \Theta_s \delta - U g_s(\Lambda) U^T \Theta_s \delta \right) \\
		&= \Psi \Theta \Psi^{-1} \delta + \sum_{s} U \left( g_s(\Lambda + \Delta\Lambda) - g_s(\Lambda) \right) U^T \Theta_s \delta ,
	\end{aligned}   
\end{equation}
where \(U\) is the feature vector matrix that represents the graph structure in the wavelet domain. \(g_s(\Lambda)\) represents the wavelet function evaluated at the original eigenvalue matrix $\Delta \Lambda$. \(g_s(\Lambda + \Delta\Lambda)\) represents the wavelet function evaluated at the perturbed eigenvalue matrix. 
\(\Theta_s\) is the corresponding learnable weight matrix for the \(s\)-th wavelet function.

Then we can use the triangle inequality:
\begin{equation}
	\begin{aligned}
	\| \sum_{s} U (g_s(\Lambda + \Delta\Lambda) - g_s(\Lambda)) U^T \Theta_s \delta \| \leq \\
	 \sum_{s} \|U\| \|g_s(\Lambda + \Delta\Lambda) -
	g_s(\Lambda)\| \|U^T\| \| \Theta_s \| \| \delta\|.
\end{aligned}
\end{equation}

Combining both parts, we obtain:
\begin{equation}
	\begin{aligned}
		\|F_{GWC}(X + \delta) - F_{GWC}(X)\| \leq \|\Psi\| \|\Theta\| \|\Psi^{-1}\| \|\delta\| +\\
		 \sum_{s} \|U\| \|g_s(\Lambda + \Delta\Lambda) - g_s(\Lambda)\| \|U^T\| \|\Theta_s\| \|\delta\|.
	\end{aligned}
\end{equation}
Here, we derive the upper bound for the stability of GWC.

%For the first term of equation \ref{sta1}, we need to analysis the behavior of $ g_s(\Lambda + \Delta\Lambda) - g_s(\Lambda)$. Note that \( g_s \) is a continuously differentiable function, so we can use a Taylor expansion to approximate it:
%\begin{equation}
% g_s(\Lambda + \Delta\Lambda) - g_s(\Lambda) \approx 	g_s'(\Lambda) \Delta\Lambda , 
%\end{equation}
%where $g_s'$ is the derivative of \( g_s \) at \( \Lambda \). Notice that $\Delta\Lambda = l_0 \delta$, then the approximate equation is:
%\begin{equation}  
%	\begin{aligned}  
%		\|GWC(X + \delta) - GWC(X)\| \leq&   
%		\sum_{s} \| U\| \|g_s(\Lambda + \Delta\Lambda) \| \\
%		&+ \| \Psi\| \| \Theta \| \| \Psi^{-1} \| \| \delta\| \\
%		&+ g_s(\Lambda) \| \|U^T\| \|\Theta_s\| \|\delta\|.  
%	\end{aligned}  
%\end{equation}

Notice that it is linearly related to the perturbation \(\delta\), thereby proving the stability of GWC.
\subsubsection{Stability of Spectral Pooling Layer}
Following a similar approach, we calculate the upper bound for the stability of spectral pooling layer.

Based on the definition of the spectral pooling layer in the original text, we note:
\begin{equation}
	F_{sp}(X) = S^T X S.
\end{equation}
Let \( \delta\) be a small perturbation on the graph embedding matrix \( X \). We can express the perturbed function as follows:
\begin{equation}
F_{sp}(X + \delta) = S^T (X + \delta) S
\end{equation}
Expanding this expression gives:
\begin{equation}
	F_{sp}(X + \delta) = S^T X S + S^T \delta S
\end{equation}
The change in output due to perturbations is given by:
\begin{equation}
	F_{sp}(X+\delta)- F_{sp}(X) = S^T \delta S
\end{equation}
That is to say,
\begin{equation}
	\|F_{sp}(X+\delta)- F_{sp}(X)\| = \|S^T\|\cdot\|S\| \cdot \|\delta\|
\end{equation}
It is noted that the difference is also a linear function of \(\delta\), indicating that spectral pooling layer is also stable.

\end{document}